\documentclass[letterpaper, 10 pt, journal, twoside]{IEEETran}
\usepackage{graphics} 
\usepackage{epsfig} 
\usepackage{times} 
\usepackage{amsmath} 
\usepackage{amssymb}  
\usepackage{amsthm}
\usepackage{siunitx}
\usepackage{url}
\usepackage{diagbox}
\usepackage{verbatim}
\usepackage{booktabs}
\usepackage{color}
\usepackage{bm}
\usepackage{hyperref}
\usepackage[ruled,vlined]{algorithm2e} 
\usepackage{graphicx} 
\usepackage{multirow}
\usepackage{upgreek}
\usepackage{color}
\usepackage{float}
\usepackage{balance}

\usepackage{pifont}

\def\BibTeX{{\rm B\kern-.05em{\sc i\kern-.025em b}\kern-.08em
		T\kern-.1667em\lower.7ex\hbox{E}\kern-.125emX}}
\newcommand{\orcid}[1]{\href{https://orcid.org/#1}{\includegraphics[width=10pt]{orcid128.png}}}

\begin{document}

\title{\LARGE \bf L-PR: Exploiting LiDAR Fiducial Marker for Unordered Low Overlap Multiview Point Cloud Registration }

\author{Yibo Liu, Jinjun Shan, Amaldev Haridevan, Shuo Zhang

 	\thanks{This work was supported by NSERC Alliance Grant ALLRP 555847-20 and Mitacs Accelerate Grant IT26108. (Corresponding author: Jinjun Shan.)}
	\thanks{Yibo Liu, Jinjun Shan, Amaldev Haridevan, and Shuo Zhang are with the Department of Earth and Space Science, Lassonde school of Engineering, York University, Toronto, Ontario, M3J1P3, Canada. email:\{yorklyb,jjshan,amaldev,shuoy\}@yorku.ca}}
	
\maketitle

\begin{abstract}
Point cloud registration is a prerequisite for many applications in computer vision and robotics. 
Most existing methods focus on the registration of point clouds with high overlap. While some learning-based methods address low overlap cases, they struggle in out-of-distribution scenarios with extremely low overlap ratios.
This paper introduces a novel framework dubbed L-PR, designed to register unordered low overlap multiview point clouds leveraging LiDAR fiducial markers.
We refer to them as LiDAR fiducial markers, but they are the same as the popular AprilTag and ArUco markers—thin sheets of paper that do not affect the 3D geometry of the environment.
We first propose an improved adaptive threshold marker detection method to provide robust detection results when the viewpoints among point clouds change dramatically.
Then, we formulate the unordered multiview point cloud registration problem as a maximum a-posteriori (MAP) problem and develop a framework consisting of two levels of graphs to address it.
The first-level graph, constructed as a weighted graph, is designed to efficiently and optimally infer initial values of scan poses from the unordered set.
The second-level graph is constructed as a factor graph. By globally optimizing the variables on the graph, including scan poses, marker poses, and marker corner positions, we tackle the MAP problem.
We perform both qualitative and quantitative experiments to demonstrate that the proposed method outperforms previous state-of-the-art (SOTA) methods in addressing challenging low overlap cases. Specifically, the proposed method can serve as a convenient, efficient, and low-cost tool for applications such as 3D asset collection from sparse scans, training data collection in unseen scenes, reconstruction of degraded scenes, and merging large-scale low overlap 3D maps, which existing methods struggle with. We also collect a new dataset named Livox-3DMatch using L-PR and incorporate it into the training of the SOTA learning-based methods, which brings evident improvements for them across various benchmarks. We release the open-source code implementation and datasets at https://github.com/yorklyb/L-PR. 
\end{abstract}

\section{Introduction}
\IEEEPARstart{P}{oint} cloud registration is a fundamental need in computer vision \cite{sghr,pre} and robotics \cite{teaser,kiss,lloam}. Most existing point registration methods \cite{teaser,3dmatch,se3et,geotransformer} focus on pairwise registration of two point clouds with high overlap, which are not robust in practical applications \cite{pre}. Some recent learning-based research proposes multiview registration methods \cite{sghr,mdgd} and studies low overlap cases \cite{pre,sghr,mdgd}. Despite their promising performance on benchmarks, the generalization of learning-based methods to unseen scenarios (\textit{i.e.}, out-of-distribution cases of training data) remains problematic. Moreover, both geometry-based methods \cite{teaser,kiss,lloam,random} and learning-based methods \cite{sghr,pre} rely on shared geometric features between point clouds to align them. Hence, these methods struggle in scenes with scarce shared features, such as extremely low overlap cases and degraded scenes \cite{degradation}. Consequently, the existing methods \cite{mdgd,sghr,se3et,geotransformer,teaser} are unsuitable for tasks such as capturing the complete 3D shape of a novel object from a sparse set of scans taken from dramatically different viewpoints, gathering data from unseen scenarios for training, reconstructing degraded scenes, or merging large-scale 3D maps with low overlap.
\par
Reviewing the domain of 2D images, it is found that fiducial markers, such as AprilTag \cite{ap3} and ArUco \cite{aruco}, have been widely used to tackle challenging situations, such as dramatic viewpoint changes (extremely low overlap) between frames and textureless environments \cite{munoz2018,munoz2019,qingdao}. The visual fiducial markers have rich patterns that are robust for detection and matching, thanks to elaborately designed encoding-decoding algorithms \cite{ap3,aruco}. They pose as thin sheets of paper that can be attached to surfaces without disrupting the 3D environment. 
In contrast, fiducial objects developed for LiDARs, such as calibration boards \cite{cal,cal2,a4}, lack encoding-decoding pattern design and often take the form of large boards mounted on tripods, thus affecting the 3D environment.
LiDARTag \cite{lt} is the first work to develop a fiducial marker system akin to AprilTag \cite{ap3} for LiDARs. While its patterns are compatible with AprilTag, the marker itself is still an additional object placed on a tripod. Recently, the intensity image-based LiDAR fiducial marker system (IFM) \cite{iilfm} proposes to utilize intensity and range images to detect markers. Thus, IFM is compatible with popular visual fiducial markers \cite{ap3,aruco} and inherits the virtue of not affecting the 3D environment. However, the marker detection of IFM is not robust when the viewpoints change in a wide range \cite{iilfm}, which hinders its application to in-the-wild multiview point cloud registration. Although the development of LiDAR fiducial markers \cite{lt,iilfm} has advanced, their applications are limited to calibration \cite{lt2} and AR \cite{iilfm}. The utilization of LiDAR fiducial markers in multiview point cloud registration remains an open problem. \par
Furthermore, training data is crucial for the development of learning-based point cloud registration methods \cite{mdgd,sghr}, but collecting it is time-consuming and expensive. For instance, in previous dataset collections \cite{3dmatch,eth,scan}, external sensors such as cameras, IMUs, and GPS are employed to obtain ground truth poses among point clouds. This requires careful multi-sensor calibration and synchronization, which can be time-consuming and labor-intensive.
\begin{figure*}[t]
	\centering
	\includegraphics[width=15.0cm]{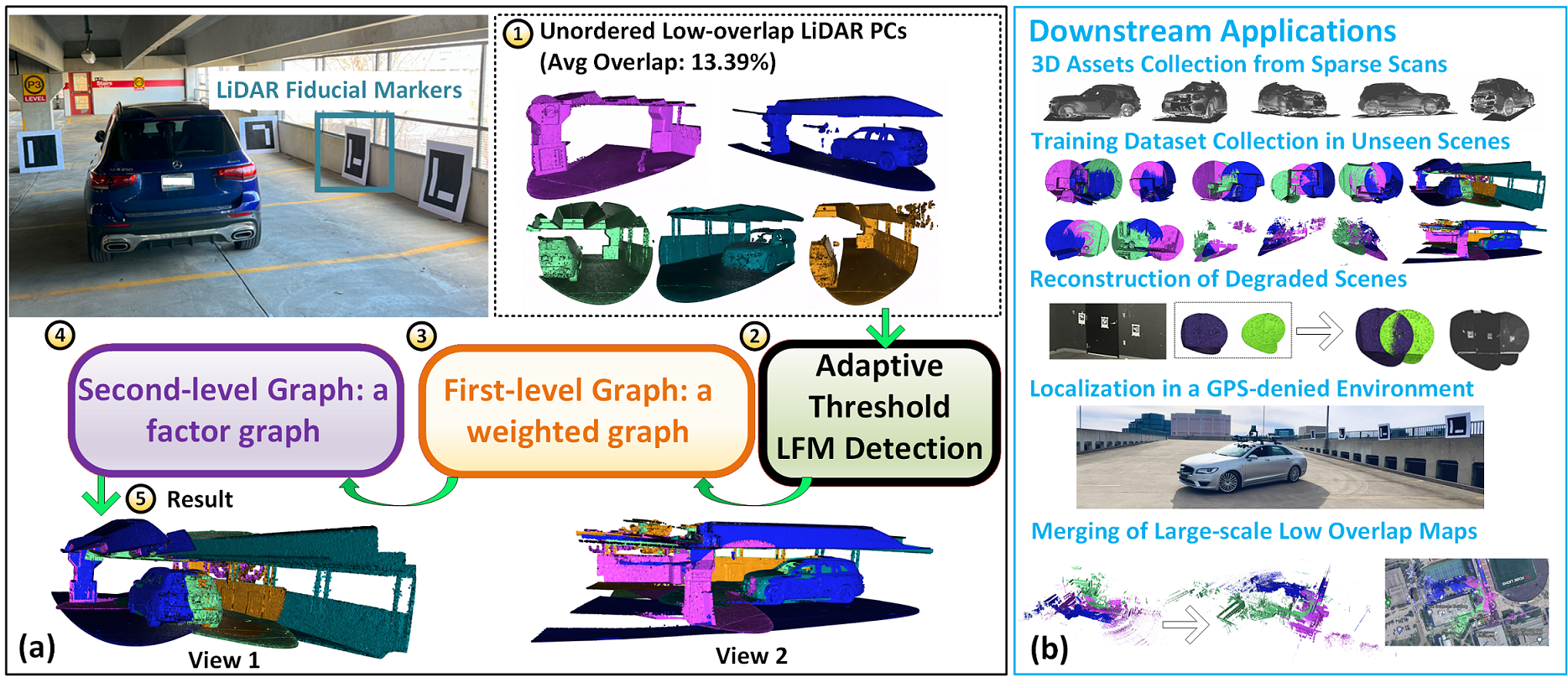}
	
	\caption{(a): \textbf{Overview of L-PR.} Given {\small \ding{172} an unordered set of unaligned, low overlap 3D point clouds}, we aim to register them into {\small \ding{176} a complete point cloud} utilizing LiDAR fiducial markers (thin sheets of paper/board attached to other planes). This is achieved through the proposed {\small \ding{173}} adaptive threshold marker detection method and the {\small \ding{174}}{\small \ding{175}} two-level graphs. The {\small \ding{174}} first-level graph handles the unordered set of point clouds and calculates initial values. The {\small \ding{175}} second-level graph registers the point clouds by finding the optimal solution to a maximum a-posteriori problem. (b): The proposed method can serve as a convenient, efficient, and low-cost tool for diverse applications, including 3D asset collection from sparse observations,  collecting training data in novel scenes, reconstruction of degraded scenes, localization in GPS-denied environments, and merging large-scale low overlap 3D maps.
    }
\label{mov}
\end{figure*}
In this paper, we develop a novel framework, dubbed L-PR (see Fig. \ref{mov}(a)), to exploit LiDAR fiducial markers for unordered low overlap multiview point cloud registration. As seen, the LiDAR fiducial markers are thin sheets of board/paper attached to other surfaces without impacting the environment. Specifically, the LiDAR fiducial markers can be printed on paper using standard printers, offering a low-cost, easy-to-use, and reliable alternative compared to previous external sensors. Given an unordered set of low overlap point clouds, the proposed L-PR efficiently registers them in a global frame. To address the unstable marker detection of IFM \cite{iilfm} in the wild, we first develop an adaptive threshold marker detection method that is robust to viewpoint switching. Then, we formulate the multiview point cloud registration problem as a maximum a-posteriori (MAP) problem and develop two-level graphs to tackle it. 
The first-level graph, constructed as a weighted graph, is designed to efficiently process the unordered set of point clouds and calculate the initial values of poses among point clouds. 
In particular, the weights represent the pose estimation error of each marker observation, and we obtain the optimal solutions of initial values by finding the shortest path from an anchor scan to each non-anchor scan on the graph.
Given the initial values, we design the second-level graph, a factor graph, to resolve the MAP problem by globally optimizing the poses of point clouds, markers, and positions of marker corners. We conduct both qualitative and quantitative experiments to demonstrate the superiority of the proposed method over competitors \cite{mdgd,sghr,se3et,geotransformer,teaser,qingdao,kiss}. Especially, the proposed L-PR is robust to any unseen scenarios with extremely low overlap, making it a convenient, efficient, and low-cost tool for diverse applications that pose significant challenges to existing methods. As shown in Fig. \ref{mov}(b), these applications include 3D asset \cite{cd,vqadiff,learning} collection from sparse observations, training data collection for learning-based models \cite{mdgd,sghr} in unseen scenes, reconstruction of degraded scenes, localization in GPS-denied environments, and merging large-scale low overlap 3D maps.

We release the code implementation, datasets, and trained model weights at https://github.com/yorklyb/L-PR.

The \textbf{contributions} of this work are as follows:
\begin{itemize}
    
    \item We develop an improved LiDAR fiducial marker detection algorithm which is robust to viewpoint changes.
	\item We design a novel framework, named \textbf{L-PR}, for unordered low overlap multiview point cloud registration leveraging LiDAR fiducial markers. Due to the robustness of the proposed L-PR to novel scenes with extremely low overlap, L-PR can serve diverse applications that existing methods struggle with.
    \item We collect a new training dataset called \textbf{Livox-3DMatch} using the proposed L-PR. Livox-3DMatch augments the original 3DMatch training data \cite{3dmatch} from 14,400 pairs to 17,700 pairs (a \textbf{22.91\%} increase). By training on this augmented dataset, the performance of the SOTA learning-based methods \cite{mdgd,sghr} is boosted across various benchmarks. SGHR \cite{sghr} is improved by \textbf{2.90\%} on 3DMatch \cite{3dmatch}, \textbf{4.29\%} on ETH \cite{eth}, and \textbf{22.72\%} (translation) / \textbf{11.19\%} (rotation) on ScanNet \cite{scan}. MDGD \cite{mdgd} is improved by \textbf{1.71\%} on 3DMatch \cite{3dmatch}, \textbf{2.89\%} on ETH \cite{eth}, and \textbf{22.45\%} (translation) / \textbf{7.80\%} (rotation) on ScanNet \cite{scan}.

\end{itemize}

\section{Related Works}

\subsection{LiDAR Fiducial Markers}
Given the success of Structure from Motion \cite{munoz2018,qingdao} and visual Simultaneous Localization and Mapping (SLAM) \cite{munoz2019,tagslam,shuo3} approaches based on visual fiducial markers \cite{ap3,aruco}, why do we still need to exploit LiDAR and LiDAR fiducial markers? There are three reasons: (1) Unlike image-based fiducial marker detection, which is sensitive to ambient light, LiDAR fiducial marker detection is robust to unideal illumination conditions \cite{lt,iilfm}, such as purely dark or overexposed environments. (2) Unlike image-based fiducial marker detection \cite{yibo1,yibo2,liao}, which suffers from rotational ambiguity \cite{munoz2018,shuo2,qingdao,munoz2019} and requires abandoning ambiguous measurements, LiDAR fiducial marker pose estimation is not affected by rotational ambiguity \cite{lt,iilfm} and each marker observation can be utilized. (3) Data captured by LiDARs is irreplaceable by data acquired from cameras due to the unique modality of LiDAR point cloud \cite{pcgen}.

\begin{figure}[hpb]
		\centering
		\includegraphics[width=3.3in]{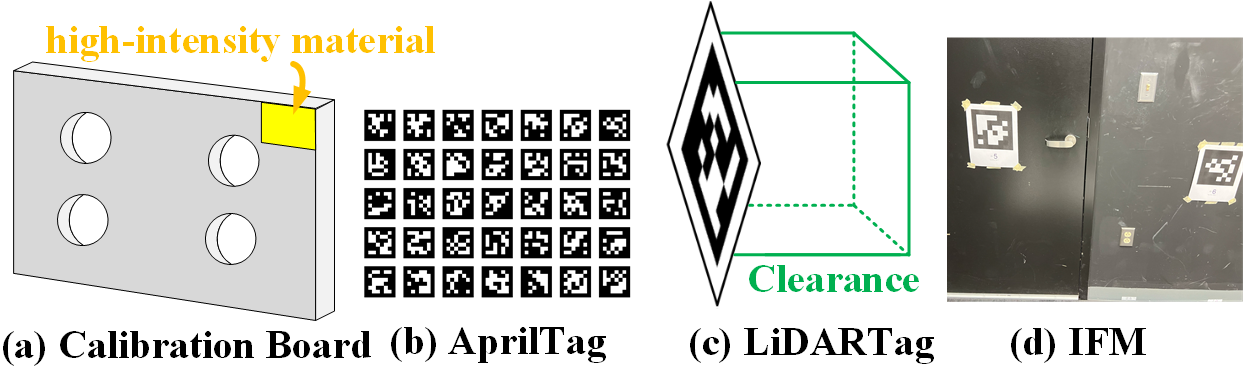}
		\caption{(a) Typical calibration board: forms a thick board with holes and/or regions with high-intensity materials. It impacts the 3D environment and does not support patterns generated by an encoding-decoding algorithm. (b) AprilTag \cite{ap3}: has rich patterns generated from the encoding-decoding algorithm and is thus robust to false positives/negatives. It has no effect on the 3D environment but is designed for cameras. (c) LiDARTag \cite{lt}: has patterns compatible with AprilTag \cite{ap3} but still forms an additional 3D object. (d) IFM \cite{iilfm}: has patterns compatible with AprilTag \cite{ap3} and ArUco \cite{aruco}. It has no impact on the 3D environment.} \label{tags}
\end{figure}

Most LiDAR fiducial objects \cite{cal,cal2,a4} are designed for calibration purposes.
Fig. \ref{tags}(a) shows a typical calibration board, a thick board with holes and/or regions covered by high-intensity materials. 
A calibration board is commonly placed on a tripod for spatial distinction, making it an additional 3D object that affects the environment.
Moreover, while a calibration board can provide fiducials (holes, corners, and high-intensity regions) and it is feasible to assign specific indexes to the fiducials, a calibration board is fundamentally different from fiducial marker systems \cite{ap3,aruco}.
The core of fiducial marker systems \cite{ap3,aruco} lies in the elaborately designed encoding-decoding algorithms, leveraging which various complex patterns are generated as depicted in Fig. \ref{tags}(b). In addition to supporting a larger number of unique IDs, the patterns of fiducial markers are robust against false positive/negative issues.
Recently, utilizing the encoding-decoding algorithm of AprilTag 3 \cite{ap3}, the first fiducial marker system for LiDARs, named LiDARTag, is introduced in \cite{lt}. Unfortunately, as presented in Fig. \ref{tags}(c), a LiDARTag is still an additional 3D object impacting the 3D environment. This is undesirable when collecting training data for learning-based point cloud registration methods.
IFM \cite{iilfm} develops a LiDAR fiducial marker system that can be integrated with various visual fiducial marker systems \cite{ap3,aruco}.
Furthermore, as seen in Fig. \ref{tags}(d), the usage/placement of IFM is as convenient as the visual fiducial marker. 
Nevertheless, the marker detection of IFM is unstable when the viewpoints change dramatically among scans due to the use of a constant threshold to process the intensity image \cite{iilfm}. 
We develop an improved adaptive threshold marker detection method to address this problem because robust marker detection is the prerequisite for exploiting markers in point cloud registration. 

\subsection{Point Cloud Registration}
Geometry-based point cloud registration methods, such as variants of iterative closest point methods \cite{kiss,lloam}, RANSAC \cite{random}, and Teaser++ \cite{teaser}, mainly focus on designing efficient and robust algorithms for describing geometry and extracting geometric features (\textit{e.g.}, points, lines, and surfaces/normals) to find correspondence between point clouds and align them. In learning-based point cloud registration methods, neural networks are designed to learn features representing the geometry \cite{mdgd,sghr,tim1,tim3,features,multiview,3dmatch,se3et,geotransformer}, and then learn to utilize these features for registering the point clouds. In addition, compared to pairwise registration methods \cite{se3et, geotransformer, teaser} that only learn to align two point clouds, multiview point cloud registration methods, such as MDGD \cite{mdgd} and SGHR \cite{sghr}, also learn to refine the pose graph constructed with the pairwise registration results.
While these methods demonstrate remarkable performance on benchmarks \cite{3dmatch,eth,scan}, they still face challenges in handling degraded scenes, such as repetitive structures and textureless environments. Moreover, most existing methods \cite{tim2,kiss,lloam,features,3dmatch} only apply to high overlap scenarios. Despite the existence of some research \cite{sghr,pre} on low overlap cases, their generalization to unseen scenes is limited. Inspired by the utilization of visual fiducial markers in camera-based applications \cite{munoz2018,qingdao, munoz2019,tagslam} to handle degraded scenes and low overlap cases, we propose exploiting LiDAR fiducial markers to enhance multiview point cloud registration. \par
\section{Preliminaries}
\subsection{Three-dimensional Transformation} \label{threedimen}
Suppose that $\mathbf{p}_{a} = [x,y,z]^{T}$ is a 3D point expressed in the coordinate system $\{a\}$. To express the point in another coordinate system $\{b\}$ as $\mathbf{p}_{b} = [x^{\prime},y^{\prime},z^{\prime}]^{T}$, translation and rotation transformations need be applied to $\mathbf{p}_{a}$. In particular, the extrinsic matrix, $\mathbf{T}\in\mathbb{R}^{4\times4}$, is employed to describe the 6-DOF pose:

\begin{equation}
\mathbf{T} = \left[\begin{array}{cc}
\mathbf{R} & \mathbf{t} \\
\mathbf{0}^{1 \times 3} & 1 \label{transform}
\end{array}\right],
\end{equation} 
where $\mathbf{t}\in\mathbb{R}^{3\times1}$ denotes the translation vector and $\mathbf{R}\in \mathbb{R}^{3\times3}$ denotes the rotation matrix. Afterwards, $\mathbf{p}_{b}$ is obtained as follows:
\begin{equation}
\left[\begin{array}{c}
\mathbf{p}_b \\
1
\end{array}\right]=\mathbf{T}\left[\begin{array}{c}
\mathbf{p}_a \\
1
\end{array}\right]. \label{dot}
\end{equation} \par
The operator ($\cdot$) is introduced to simplify Eq. (\ref{dot}) as:
\begin{equation}
\mathbf{p}_b=\mathbf{T} \cdot \mathbf{p}_a. 
\end{equation}

\subsection{Lie Group and Lie Algebra} \label{lie}
The rotation matrix $\mathbf{R}\in SO(3)$, which is the special orthogonal group representing rotations \cite{barfoot}:
\begin{equation}
SO(3) = \{ \mathbf{R}\in\mathbb{R}^{3\times3} | \ \mathbf{R}\mathbf{R}^{T}=\mathbf{1},det(\mathbf{R}) =1 \}.
\end{equation} \par
The pose matrix $\mathbf{T} \in SE(3)$, which is the special Euclidean group representing poses \cite{barfoot}:
\begin{equation}
SE(3) = \{ \mathbf{T} = \left[\begin{array}{cc}
\mathbf{R} & \mathbf{t} \\
0 & 1 
\end{array}\right] \in \mathbb{R}^{4 \times 4} | \ \mathbf{R} \in SO(3), \mathbf{t}\in \mathbb{R}^{3 \times 1}\}.
\end{equation} \par
$SO(3)$ and $SE(3)$ are two specific Lie groups. Every matrix Lie group is associated with a Lie algebra \cite{barfoot}. The Lie algebra $\mathfrak{so}(\mathrm{3})$, associated with $SO(3)$, is defined as follows:
\begin{equation}
\begin{aligned}
& SO(3) \rightarrow \mathfrak{so(\mathrm{3})}: \\
& \log(\mathbf{R}) = \frac{\theta}{2\mathrm{sin}(\theta)}(\mathbf{R}-\mathbf{R}^{T}),  \\ 
& \xi = \log(\mathbf{R})_{\vee}, 
\end{aligned}
\end{equation}
where $\theta= \arccos \frac{1}{2}(Trace(\mathbf{R})-1)$. $\log(\cdot)$ is the matrix logarithm and $\xi \in \mathbb{R}^{3\times1}$ is the Lie algebra coordinates. $\vee$ is the \textit{vee} map operator that finds the unique vector $\xi \in \mathbb{R}^{3\times1}$ corresponding to a given skew-symmetric matrix $\log(\mathbf{R})\in \mathbb{R}^{3 \times 3}$ \cite{barfoot,tagslam}.
\section{Methodology}
\subsection{Adaptive Threshold LiDAR Fiducial Marker Detection} \label{adaptive}
\noindent\textbf{Vanilla IFM.}
Given a 3D point cloud, the vanilla IFM \cite{iilfm} generates a 2D intensity image by mapping each 3D point to a pixel on an image plane:
\begin{equation}	
	u = \lceil \frac{\theta}{\alpha_{a}}\rfloor + u_{o},\; v = \lceil \frac{\phi}{\alpha_{i}}\rfloor + v_{o},
\label{pro}
\end{equation}
where $[u,v]^{T}$ denote the image coordinates of the projected pixel. $[\theta,\phi,r]^{T}$ denote the spherical coordinates of the 3D point. $\alpha_{a}$ and $\alpha_{i}$ represent the angular resolutions in $u$ (azimuth) and $v$ (inclination) directions, respectively. $u_{o}$ and $v_{o}$ are the offsets. The pixel value $\mathbf{I}(u,v)$ is decided by the corresponding intensity value through the color map. The ranging information is also saved for each pixel. 

In the real world, the intensity values of the same surface of an object could vary, leading to noise in the rendered image. To denoise the intensity image, each pixel is binarized with a threshold $\lambda$. In \cite{iilfm}, $\lambda$ is set as a constant. Then, the embedded 2D marker detector \cite{ap3,aruco} is applied to localize the fiducials on the binarized image. Using the ranging information, the detected 2D fiducials are projected back into 3D space via the reverse of Eq. (\ref{pro}) and become 3D fiducials. \par
\begin{figure}[thb]
	\centering
	\includegraphics[width=3.3in]{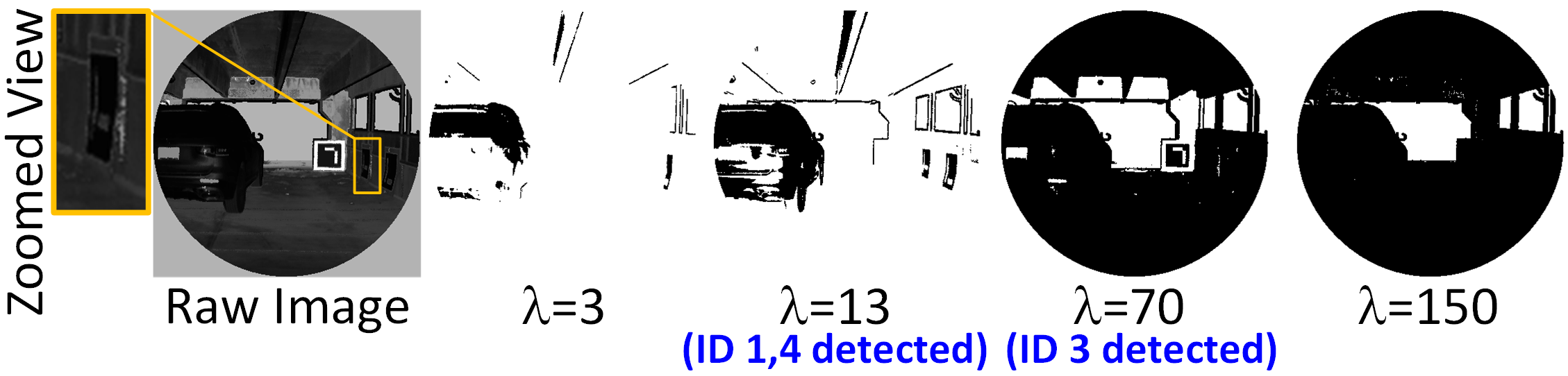}

	\caption{The raw intensity image binarized with different threshold values.}
	\label{threshold}
\end{figure}

\noindent\textbf{Adaptive Threshold Method.}
As aforementioned, binarization is applied to the raw intensity image due to the imaging noise (see the zoomed view of Fig. \ref{threshold}). As seen, the effect is determined by the threshold $\lambda$.
In \cite{iilfm}, it is found that $\lambda$ can be selected as a constant if the viewpoint does not change drastically. However, with significant changes in the scene, the value of $\lambda$ needs adjustment. 
As an example, consider the case shown in Fig. \ref{threshold}, where three markers are placed. Markers ID 1 and 4 are detectable when $\lambda$=13, while marker ID 3 is detectable when $\lambda$=70. Thus, $\lambda$=13 and $\lambda$=70 are the optimal thresholds compared to other values for markers ID 1 and 4 and marker ID 3, respectively, denoted by $\lambda^{*}$. \par
Therefore, unless an appropriate $\lambda$ is carefully selected for each scan, and even for each marker, the LiDAR fiducial marker detection will fail.
Unlike \cite{iilfm}, which determines $\lambda^{*}$ based on experience or experimentation, we develop an algorithm to automatically seek $\lambda^{*}$.
%
Given that our focus is on marker detection rather than image denoising, we design \textbf{Algorithm \ref{algo1}} that utilizes the detection result as the feedback to search for $\lambda^{*}$ automatically. In particular, the core of the algorithm is to maximize the length of a memory queue for saving detected markers (denoted by $Q$) by gradually increasing $\lambda$. Namely, we aim to detect as many markers as possible by finding the optimal threshold for each marker in the scene. After applying \textbf{Algorithm \ref{algo1}}, the 2D fiducials located on the image binarized with $\lambda^{*}$ are projected into 3D fiducials using the subsequent steps of IFM \cite{iilfm}. Our algorithm addresses the problem of a constant threshold being inapplicable to all scans and markers, especially when LiDAR moves in the wild and scenes undergo significant changes.
\begin{algorithm}[h]
    \KwIn{Raw intensity image, $\mathbf{I}$}
    \KwOut{The optimal threshold, $\lambda^{*}$.}
    \label{algo1}
    \textbf{Initialize parameters:} Search scope, $S$. step size, $\delta$. 
    queue for saving detected markers, $Q=[ \ ]$. length of $Q$, $Q_l=0$.
    temporary queue, $Q_{temp}=[\ ]$. length of  $Q_{temp}=[\ ]$,  $Q_{temp,l}=0$. the optimal threshold, $\lambda^{*}=0$. search step, $i=0$.\\
    Define the binarization operation as $\Psi(\mathbf{I},\lambda)$. \\ Define the marker detector operation as $\Gamma(\mathbf{I})$. \\
    \While{$i < S$}{
    $\lambda = \delta\times i$\\
        $\mathbf{I} = \Psi(\mathbf{I},\lambda)$\\
         $\Gamma(\mathbf{I}) \rightarrow Q_{temp}$\\
        \If{$Q_{temp,l} \geq Q_{l}$}{
        \For{marker \textbf{in} $Q_{temp}$}{
        \If{marker \textbf{not in} $Q$}{append \textit{marker} to $Q$}
        }
        $\lambda^{*}=\lambda$}
        $i = i+1$\\     
    }
    Return the image binarized with $\lambda^{*}$ and $Q$.
    \caption{ Search for the optimal threshold, $\lambda^{*}$. }
\end{algorithm}
\vspace{-0.1in}
\subsection{Problem Formulation} \label{problem}
In this section, we introduce the problem formulation. Suppose that the marker size is $l$. In the marker coordinate system $\{M\}_{j}$, we know that the 3D coordinates of the four corners of the $j$-th marker, $^{j}\mathbf{p}^{j,1},^{j}\mathbf{p}^{j,2},^{j}\mathbf{p}^{j,3},$ and $^{j}\mathbf{p}^{j,4}$, are $[-l/2,-l/2,0]^{T}, [l/2,-l/2,0]^{T}, [l/2,l/2,0]^{T},$ and $[-l/2,l/2,0]^{T}$, respectively. LiDAR fiducial marker detection returns the 3D coordinates of the corners expressed in $\{F\}_{i}$ (the local coordinate system of the $i$-th scan $f_{i}$) \cite{iilfm}. Thus, the 6-DOF transmission from $\{M\}_{j}$ to $\{F\}_{i}$, denoted as $\mathbf{T}^{j}_{i}$, can be found by solving the following least square problem:
 \begin{equation}	
	\mathbf{T}^{j,*}_{i}=\underset{\mathbf{T}^{j}_{i}}{\arg \min } \sum_{s=1}^{4}\left\| \mathbf{T}^{j}_{i} \ \cdot \ ^{j}\mathbf{p}^{j,s}-_{i}\mathbf{p}^{j,s}\right\|^{2},\label{least}
\end{equation}
where the operator ($\cdot$) is introduced in Section \ref{threedimen}. The Singular Value Decomposition (SVD) method \cite{barfoot} is employed to compute $\mathbf{T}^{j,*}_{i}$. We denote the set of measurements, including 
(1) the corner positions \textit{w.r.t.} $\{M\}_{j}$, (2) the corner positions \textit{w.r.t.} $\{F\}_{i}$, and (3) the marker poses \textit{w.r.t.} $\{F\}_{i}$, as $\mathcal{Z}$. 
To register the multiview point clouds, we need to find a globally consistent pose for every scan so that the point clouds can be transformed into a complete point cloud in the global coordinate system $\{G\}$. 
To achieve this objective, we consider the following variables in this work: (1) the poses of point clouds \textit{w.r.t.} $\{G\}$, (2) the poses of markers \textit{w.r.t.} $\{G\}$, and (3) the marker corner positions \textit{w.r.t.} $\{G\}$. We specify the set of variables as $\Theta$. Finally, a maximum a-posteriori (MAP) inference problem is formulated: given the measurements, $\mathcal{Z}$, the goal is to find the optimal variable set $\Theta^{*}$ that maximizes the posterior probability $P(\Theta \mid \mathcal{Z})$:
\begin{equation}
\
	\Theta^*=\underset{\Theta}{\arg \max } \ P(\Theta \mid \mathcal{Z}).\label{map}
\end{equation} 
\par
In the following, we design a framework consisting of two-level graphs to resolve this MAP problem. Inspired by the coarse-to-fine pipeline of SGHR \cite{sghr}, we develop a first-level graph to efficiently and exhaustively determine the relative poses among point clouds, and a second-level graph to globally optimize the variables.
\subsection{First-level Graph} \label{1level}
As aforementioned, the variables in $\Theta$ to be optimized cannot be directly obtained through the measurements in $\mathcal{Z}$. Moreover, deriving the initial values of the variables requires both efficiency and accuracy, given that the input is an unordered set of low overlap point clouds. To address this challenge, we design the first-level graph. First, we consider computing the relative pose between two point clouds. Suppose the $j$-th marker appears in the scenes of two scans $f_{i}$ and $f_{m}$, $\mathbf{T}^{j}_{i}$ and $\mathbf{T}^{j}_{m}$ can be calculated using Eq. (\ref{least}). Consequently, the relative pose between $f_{i}$ and $f_{m}$, denoted as $\mathbf{T}_{i,m}$, is available from
 \begin{equation}	
	\mathbf{T}_{i,m}=(\mathbf{T}^{j}_{i})^{-1} \mathbf{T}^{j}_{m},\label{relative}
\end{equation}
where $(\mathbf{T}^{j}_{i})^{-1}$ indicates the inverse of $\mathbf{T}^{j}_{i}$. $(\mathbf{T}^{j}_{i})^{-1}$ indicates the pose that transforms 3D points from $\{F\}_{i}$ to $\{M\}_{j}$. Although the method introduced in Eq. (\ref{relative}) is straightforward, it cannot be directly applied because the input in this work is an unordered set of point clouds that do not follow a temporal or spatial sequence.
Specifically, for scans with no shared marker observations, their relative pose has to be calculated through pose propagation among other scans. However, multiple alternative paths could exist. Even for scans that share marker observations, there could be more than one overlapped marker. Hence, to accurately and efficiently estimate relative poses among scans, it is necessary to design an algorithm to determine which scans and markers to apply Eq. (\ref{relative}). 
\par
\begin{figure}[t]
	\centering
\includegraphics[width=3.3in]{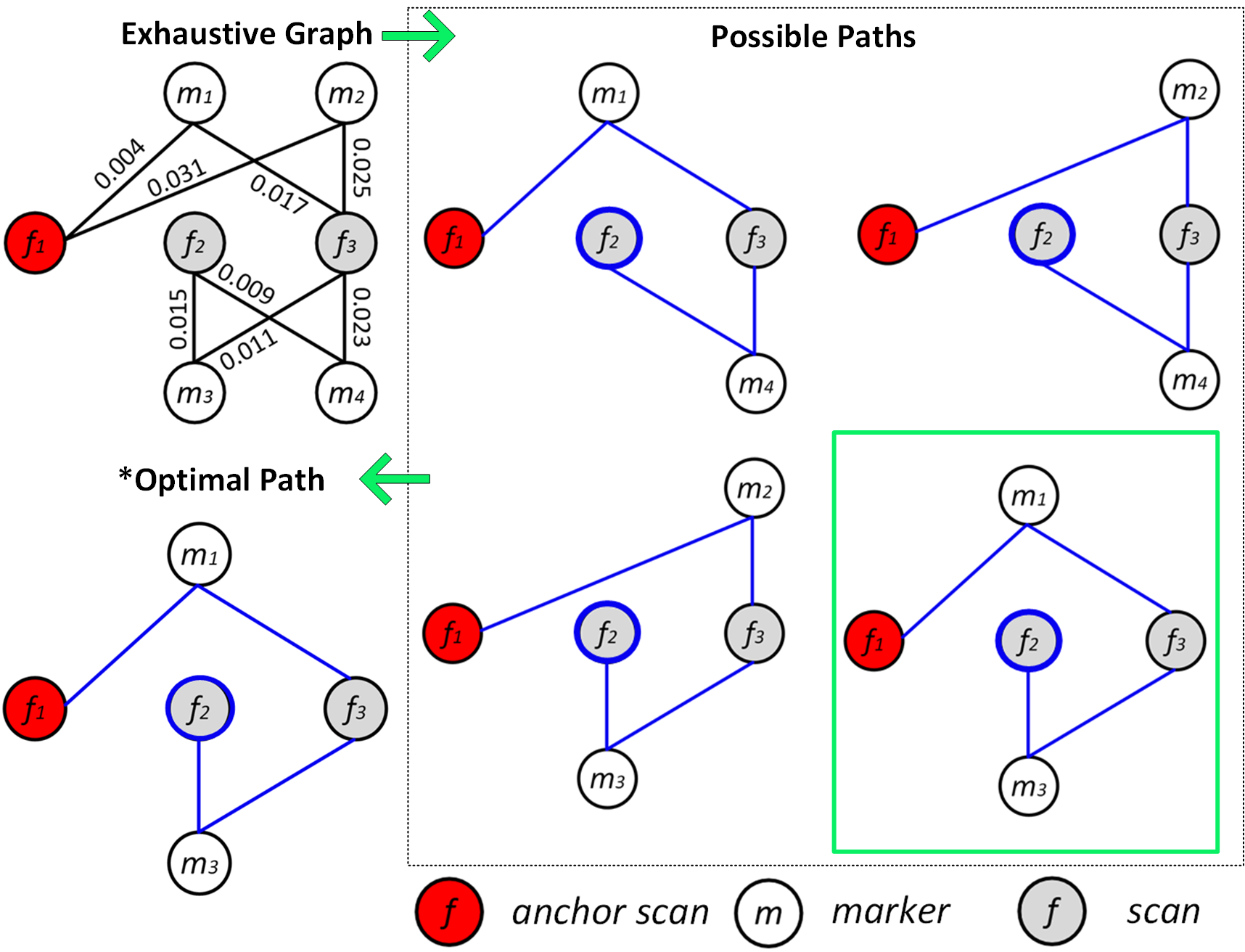}

	\caption{An illustration of the first-level graph.
    After applying the proposed adaptive marker detection to all scans, an exhaustive weighted graph is constructed, with the scans and markers as nodes and the point-to-point errors as edge weights.
    The aim is to derive the relative pose between each non-anchor scan and the anchor scan with optimal accuracy.
    However, for a given non-anchor scan, such as $f_{2}$ in this simple case, there may be multiple possible paths in the exhaustive graph leading to the anchor scan ($f_{1}$).
    Thus, Dijkstra’s algorithm \cite{dij} is employed to find the optimal path with the minimum accumulation of point-to-point errors (weights).
    }
	\label{first}
\end{figure}

The objective is to infer the relative poses with the highest possible accuracy using only the necessary low-dimensional information through a simple process. 
Thus, we construct the first-level graph, shown in Fig. \ref{first}, as a weighted graph. The construction process is as follows.
When processing the point clouds with the proposed adaptive threshold method one by one, if a marker is detected in a point cloud, we add the corresponding scan node and marker node to the graph along with a weighted edge. The edge weight is the pose estimation point-to-point error ${e}_{pp}$:
\begin{equation}	
	{e}_{pp}=\sum_{s=1}^{4}\left\| \mathbf{T}^{j,*}_{i} \ \cdot \ ^{j}\mathbf{p}^{j,s}-_{i}\mathbf{p}^{j,s}\right\|^{2}.\label{epp}
\end{equation}
\par
Eq. (\ref{epp}) indicates the substitution of the result given by SVD back into the right side of Eq. (\ref{least}). ${e}_{pp} \in \mathbb{R}^{+}$ is employed as the metric to evaluate the quality of pose estimation of the marker in the corresponding point cloud. We define the first scan as the anchor scan. The local coordinate system of the anchor scan is set as the global coordinate system. 
Namely, $\{F\}_{1}=\{G\}$. We only consider the relative poses between the anchor scan and each non-anchor scan. Although there may be multiple paths from the given scan to the anchor scan (see Fig. \ref{first}), we have already saved the information on pose estimation quality by constructing the first-level graph as a weighted graph. Therefore, we adopt Dijkstra’s algorithm \cite{dij} to obtain the shortest path. Then, we compute the relative pose along the shortest path iteratively using Eq. (\ref{relative}). Given that the relative pose computation is achieved with the lowest accumulation of $e_{pp}$, the estimation result achieves the highest possible accuracy. Moreover, the search for the optimal path to propagate poses is based merely on the one-dimensional $e_{pp}$, without incorporating any 6-DOF poses or 3D locations.
\par
In this way, the point cloud poses \textit{w.r.t.} $\{G\}$ are obtained. Since the marker detection provides the marker poses and the 3D positions of marker corners \textit{w.r.t.} the local coordinate system of corresponding scan, we can derive the initial values of the marker poses \textit{w.r.t.} $\{G\}$ and the corner positions \textit{w.r.t.} $\{G\}$ through the point cloud poses \textit{w.r.t.} $\{G\}$.

 \par

\subsection{Second-level Graph}
To address the MAP problem introduced in Eq. (\ref{map}), we construct the second-level graph as a factor graph to globally optimize the variables using the initial values obtained from the first-level graph. Specifically, we create the second-level graph in three stages.\par
\begin{figure}[t]
	\centering
	\includegraphics[width=3.3in]{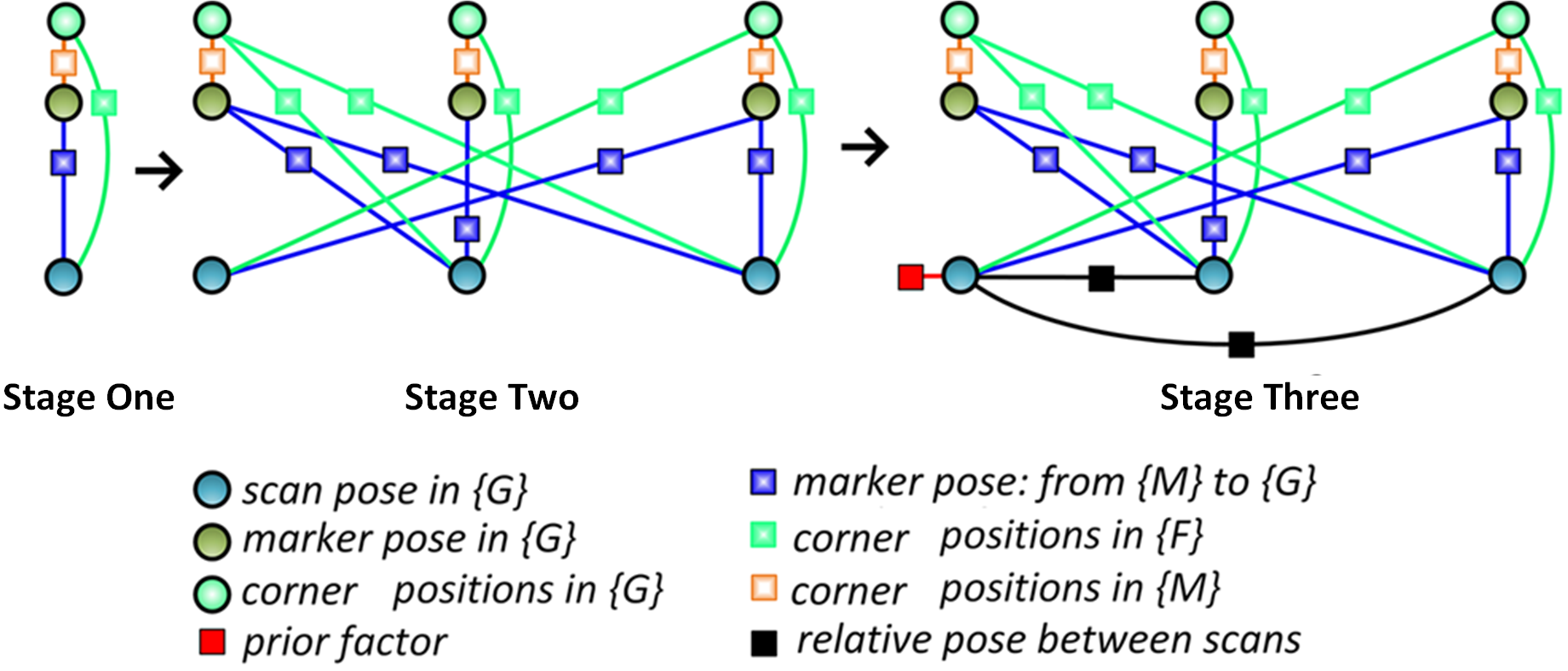}
	\caption{The procedures for formulating the second-level graph. The variable nodes are represented by circles and the factor nodes are represented by squares. In Stage One, when a marker is detected in a scan, six types of nodes are added to the graph, including (1) scan pose in $\{G\}$, (2) marker pose in   $\{G\}$, (3) corner positions in $\{G\}$, (4) marker pose from $\{M\}$ to $\{G\}$, (5) corner positions in $\{F\}$, and (6) corner positions in $\{M\}$, along with their corresponding edges. In Stage Two, all the marker detection results are traversed, and the operation from Stage One is repeated for each detected marker.  In Stage Three, a prior factor connecting the anchor scan is added, along with factors representing the relative poses between the anchor scan and non-anchor scans. }\label{second}

\end{figure}

\textbf{Stage One.} Suppose that the $j$-th marker is detected in the $i$-th scan, the following six types of nodes are added to the second graph. The variable nodes include: 
\begin{itemize}
    \item [(1)] Node $\mathbf{T}^{j}$, which refers to the 6-DOF pose of the $j$-th marker \textit{w.r.t.} $\{G\}$;
    \item [(2)] Node $\mathbf{T}_{i}$, which refers to the 6-DOF pose of the $i$-th scan \textit{w.r.t.} $\{G\}$;
    \item [(3)] Nodes $\{ \mathbf{p}^{j,1}, \mathbf{p}^{j,2}, \mathbf{p}^{j,3}, \mathbf{p}^{j,4} \}$, which refer to 3D coordinates of the corners of the $j$-th marker \textit{w.r.t.} $\{G\}$. 
\end{itemize}
\noindent The factor nodes include: 
\begin{itemize}
    \item [(4)] Node $\mathbf{T}^{j}_{i}$, which refers to the measurement of the 6-DOF pose of the $j$-th marker \textit{w.r.t.} $\{F\}_{i}$;
    \item [(5)] Nodes $\{ ^{j}\mathbf{p}^{j,1}, \ ^{j}\mathbf{p}^{j,2}, \ ^{j}\mathbf{p}^{j,3}, \ ^{j}\mathbf{p}^{j,4} \}$, which refer to the measurement of the 3D coordinates of the corners of the $j$-th marker \textit{w.r.t.} $\{M\}_{j}$;

    \item [(6)] Nodes $\{ _{i}\mathbf{p}^{j,1}, \ _{i}\mathbf{p}^{j,2},$\ $ _{i}\mathbf{p}^{j,3},$ $_{i}\mathbf{p}^{j,4} \}$, which refer to 3D coordinates of the corners of the $j$-th marker \textit{w.r.t.} $\{F\}_{i}$.
\end{itemize}
By adding variables representing the 3D coordinates of the corners, we can also optimize the fiducial localization results. The first stage in Fig. \ref{second} shows the added nodes and edges when a marker is detected in a scan. \par
\textbf{Stage Two.} Thereafter, we traverse all the marker detection results and conduct the operation from Stage One for each detected marker. 
After processing all the detected markers, the second-level graph becomes the one shown in Stage Two of Fig. \ref{second}. Since the operation in this stage essentially consists of repetitions of Stage One, the node types are no different from those in the previous stage. \par
\textbf{Stage Three.} Considering that the local coordinate system of the first scan, $\{F\}_{1}$, is set as the global coordinate system, $\{G\}$, we add a prior factor that connects to the first (anchor) scan node,  $\mathbf{T}^{1}$. Finally, we add factor nodes representing the relative poses between the anchor scan and each non-anchor scan. Up to this point, the factor graph is completed, as shown in Stage Three of Fig. \ref{second}. \par
Following \cite{isam2,gtsam}, since the factor graph is determined, the posterior probability $P(\Theta \mid \mathcal{Z})$ is factorized as:
\begin{equation}
P(\Theta \mid \mathcal{Z})=\prod_k P^{(k)}(\Theta),
\label{obj}
\end{equation}
where $P^{(k)}$ are the factors in the second graph. We follow the standard pipeline \cite{isam2} to model them as Gaussians:
\begin{equation}
 P^{(k)}(\Theta) \propto \exp \left(-\frac{1}{2}\left\|h_k\left(\Theta\right)\ominus z_k\right\|_{\Sigma_k}^2\right),
\end{equation}
where $h_k\left(\Theta \right)$ is the measurement function and $z_k$ is a measurement.
$\|e\|_{\Sigma}^2 \triangleq e^T \Sigma^{-1} e$ denotes the squared Mahalanobis distance with $\Sigma$ being the covariance matrix. 
Following \cite{tagslam}, if $z_k\in\mathbb{R}^{3\times1}$ is a 3D position, $\ominus$ refers to straight subtraction for elements. 
If $z_k\in SE(3)$ is a 6-DOF pose, $\ominus$ generates 6-dimensional Lie algebra (refer to Section \ref{lie}) coordinates:
\begin{equation}{\resizebox{0.88\hsize}{!}{$\mathbf{T}_{A} \ominus \mathbf{T}_{B} = [[\log(\mathrm{Rot}(\mathbf{T}_{B}^{-1}\mathbf{T}_{A}))]^{T}_{\vee}, \mathrm{Trans}(\mathbf{T}_{B}^{-1}\mathbf{T}_{A})^{T} ]^{T},
$} } \end{equation} 
where for a $\mathbf{T} \in SE(3)$, $\mathrm{Rot}(\mathbf{T})$ denotes the rotation matrix $\mathbf{R} \in SO(3)$ and $\mathrm{Trans}(\mathbf{T})$ denotes translation vector $\mathbf{t \in \mathbb{R}^{3\times1}}$. 
$\log(\cdot)$ represents the matrix logarithm. $\vee$ is the \textit{vee} map operator. The detailed introduction of $\vee$ is provided in Section \ref{lie}.  With the Gaussian modeling, the objective function in Eq. (\ref{obj}) is transformed into a least square problem by applying the negative logarithm:
\begin{equation}
\resizebox{0.88\hsize}{!}{$\arg \min _{\Theta}(-\log \prod_k P^{(k)}(\Theta))= 
\arg \min _{\Theta} \frac{1}{2} \sum_k\left\|h_k\left(\Theta\right)\ominus z_k\right\|_{\Sigma_k}^2.$}
\end{equation} \par

We utilize the Levenberg-Marquardt algorithm \cite{gtsam} to solve this problem. The acquisition of initial values is introduced in Section \ref{1level}. The noise covariance matrices are determined by the quantitative experiments conducted in \cite{iilfm}. The code implementation, based on Python and C++, is available at https://github.com/yorklyb/L-PR. \par
 \par

\section{Livox-3DMatch Dataset} \label{secfour}
\begin{figure}[ht]
	\centering
	\includegraphics[width=3.3in]{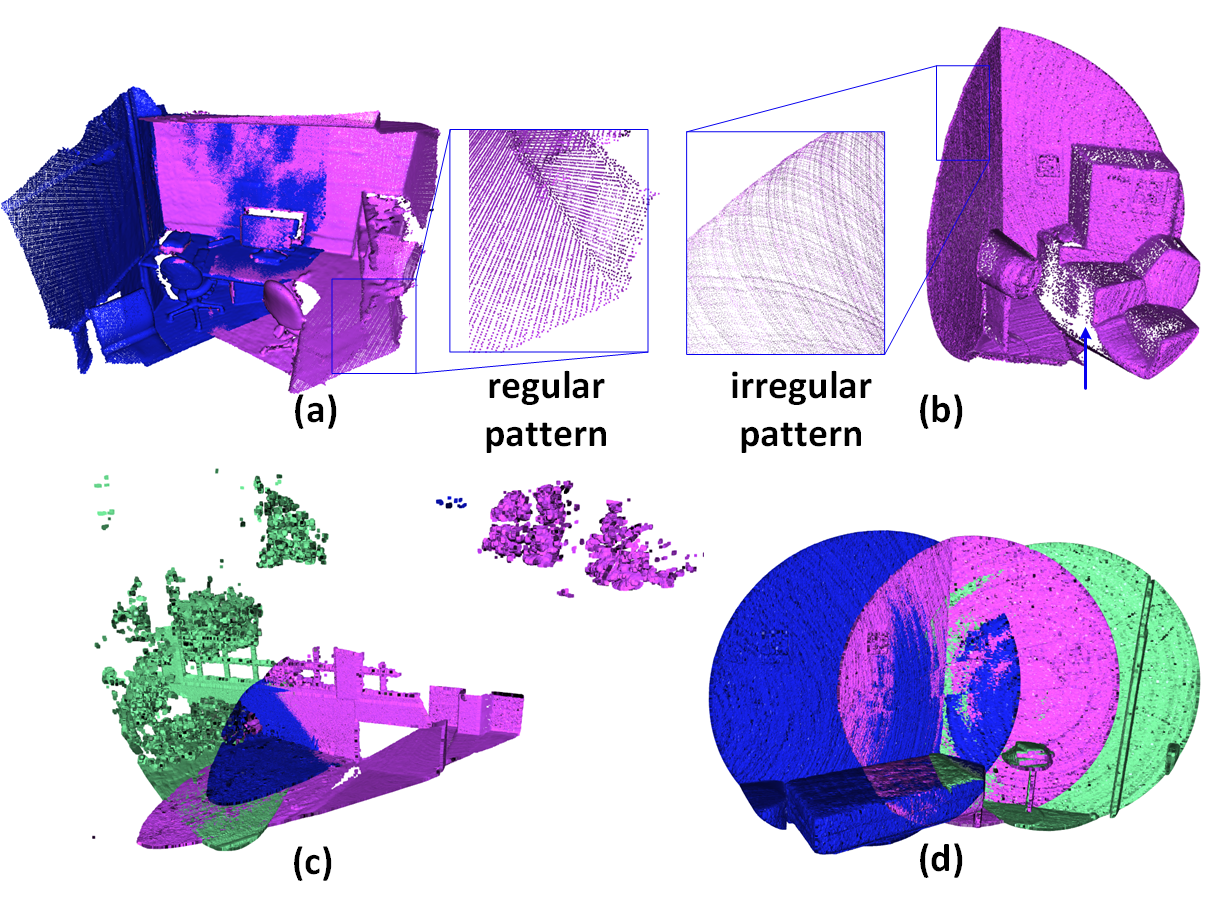}
	\caption{Comparison of 3DMatch and Livox-3DMatch. (a): A random sample from 3DMatch. The point cloud sampled by an RGB-D camera has a regular pattern and less noise. (b): A random sample from Livox-3DMatch. The Livox LiDAR point cloud has an irregular pattern and more noise. (c): An example of an outdoor scene in Livox-3DMatch. (d): A selectively sampled challenging case in which the overlapping regions are mainly planes.}
	\label{livox3d}
\vspace{-0.15in}
\end{figure}

A common approach \cite{sghr,pre} to evaluating a learning-based point cloud registration model is to train it on 3DMatch \cite{3dmatch} and test it on various benchmarks, including 3DMatch \cite{3dmatch}, ETH \cite{eth}, and ScanNet \cite{scan}. However, the 3DMatch benchmark is mainly constructed from RGB-D camera captures of indoor scenes \cite{3dmatch}. In this work, we collect a new dataset named Livox-3DMatch to enrich the training data for learning-based methods. The enrichment contains two key components. Firstly, the sensor we adopt is a Livox MID-40 LiDAR, which has different sampling patterns compared to the RGB-D camera (see Fig. \ref{livox3d}(a) and (b)). Thus, it adds point clouds with new features to the training data. Secondly, we selectively sampled some scenes that are absent or rare in 3DMatch, thereby enriching the scenes in the training data. For example, the valid depth range of an RGB-D camera is usually less than ten meters, making it unsuitable for sampling outdoor scenes. In contrast, we collect some outdoor scenes (see Fig. \ref{livox3d}(c)) for Livox-3DMatch, considering that a LiDAR can sample objects a few hundred meters away. Moreover, we collect some challenging cases (see Fig. \ref{livox3d}(d)) where the overlapping regions are mainly planes, which are rare in 3DMatch \cite{3dmatch}. A more detailed introduction to the scenes in Livox-3DMatch is given in Sections \ref{test1} and \ref{test2}. \par
In Section \ref{testadd}, we demonstrate that the proposed Livox-3DMatch can boost the performance of the SOTA learning-based methods \cite{mdgd,sghr} on various benchmarks \cite{3dmatch,eth,scan}.
\section{Experimental Results}
In this section, we report the experimental results of eight tests to demonstrate the effectiveness and applications of the proposed method from different perspectives.
\subsection{Evaluation of the Adaptive Marker Detection Method} \label{test0}
\begin{figure}[t]
	\centering
	\includegraphics[width=3.4in]{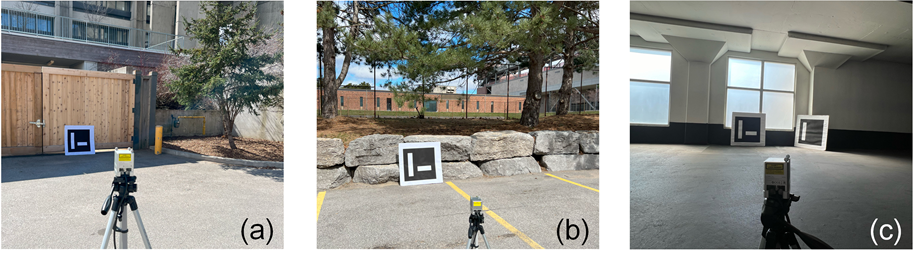}
	\caption{Setup for testing the adaptive threshold marker detection algorithm: (a) between groups of buildings, (b) in open outdoor areas, and (c) in large indoor parking lots.}
	\label{threshold2}

\end{figure}

In this test, we demonstrate the effectiveness and necessity of the proposed adaptive marker detection method.
\\
\noindent\textbf{Data.} 
We place the Livox MID-40 LiDAR and 69.2 cm $\times$ 69.2 cm ArUco marker(s) in three different scenes, as shown in Fig. \ref{threshold2}. The three scenes are: between groups of buildings, in open outdoor areas, and in large indoor parking lots. In each scene, we collect point clouds and test the proposed adaptive threshold method at different relative positions. \\
\noindent\textbf{Results and Analysis.} Table \ref{tabthreshold} presents the results, where x and y indicate the relative position of the LiDAR in the marker (ID 4) coordinate system. Specifically, in scene Fig. \ref{threshold2}(c), there are two markers (ID 4 and ID 1), so the column for $\lambda^{*}$ contains two values. As seen in the table, $\lambda^{*}$ varies significantly as the relative position changes, demonstrating the necessity of the proposed adaptive marker detection method. In addition, the results shown in Fig. \ref{threshold2}(c) also illustrate the necessity of the memory queue design, considering that a single $\lambda^{*}$ might not be applicable to all markers in the same scene. Please note that the intention of reporting the values of the optimal threshold in Table \ref{tabthreshold} is to demonstrate that a constant threshold, as adopted in \cite{iilfm}, is not applicable to this task. However, the major concern of the adaptive threshold marker detection algorithm is to detect all markers in a scene. Namely, the marker detection results saved in the memory queue are the most important output rather than the thresholds. 

\begin{table}[t]
\caption{Demonstration of the necessity of the proposed adaptive marker detection algorithm.}
\begin{center}

\resizebox{1.0\columnwidth}{!}{

\begin{tabular}{c|c|c|c | c|c|c|c| c|c|c|c}
\hline\hline
Scene & x (m) & y (m) & $\lambda^{*}$ &Scene & x (m) & y (m) & $\lambda^{*}$ &Scene & x (m) & y (m) & $\lambda^{*}$ \\ \hline
 \multirow{21}{*}{ Fig. \ref{threshold2}(a)} &0 &4   & 36  & \multirow{21}{*}{ Fig. \ref{threshold2}(b)} &0 &4   & 7 & \multirow{21}{*}{ Fig. \ref{threshold2}(c)} &4 &4   & 20/6   \\  \cline{2-4} \cline{6-8} \cline{10-12} 
 &4 &4  & 9 &  &4 &4  & 8 &  &4 &4  & 8/8 \\  \cline{2-4} \cline{6-8} \cline{10-12} 
&-4 &4   & 6 & &-4 &4  & 6 &  &-4 &4  & 6/8   \\  \cline{2-4} \cline{6-8} \cline{10-12} 
 &0 &5  & 20 &  &0 &5  & 9 &  &0 &5  & 15/10 \\  \cline{2-4} \cline{6-8} \cline{10-12} 
  &5 &5 & 8 & &5 &5  & 8&  &5 &5  & 8/12\\  \cline{2-4} \cline{6-8} \cline{10-12} 
   &-5 &5  & 7&  &-5 &5  & 9 &  &-5 &5  & 7/13 \\  \cline{2-4} \cline{6-8} \cline{10-12} 
    &0 &6  & 19 &  &0 &6  & 9 &  &0 &6  & 8/13\\  \cline{2-4} \cline{6-8} \cline{10-12} 
  &6 &6 & 10& &6 &6  & 6 &  &6 &6  & 11/11\\  \cline{2-4} \cline{6-8} \cline{10-12} 
   &-6 &6  & 11 & &-6 &6  & 8 &  &-6 &6  & 11/10 \\  \cline{2-4} \cline{6-8} \cline{10-12} 
     &0 &7  & 17 &&0 &7  & 9 &  &0 &7  & 19/20 \\  \cline{2-4} \cline{6-8} \cline{10-12} 
  &7 &7 & 8 &&7 &7  & 8 &  &7 &7  & 15/14\\  \cline{2-4} \cline{6-8} \cline{10-12} 
   &-7 &7  & 9  &&-7 &7  & 9&  &-7 &7  & 28/16  \\  \cline{2-4} \cline{6-8} \cline{10-12} 
    &0 &8  & 15  &&0 &8  & 13&  &0 &8  & 10/16 \\  \cline{2-4} \cline{6-8} \cline{10-12} 
  &8 &8 & 9 &&8 &8  & 11 &  &8 &8  & 11/17\\  \cline{2-4} \cline{6-8} \cline{10-12} 
   &-8 &8  & 10  &&-8 &8  & 15&  &-8 &8  & 12/15 \\  \cline{2-4} \cline{6-8} \cline{10-12} 
    &0 &9  & 17  &&0 &9  & 10 &  &0 &9  & 13/11\\  \cline{2-4} \cline{6-8} \cline{10-12} 
  &9 &9 & 8 &&9 &9  & 12 &  &9 &9  & 20/26\\  \cline{2-4} \cline{6-8} \cline{10-12} 
   &-9 &9  & 9  &&-9 &9  & 12  &  &-9 &9  & 16/12\\ \cline{2-4} \cline{6-8} \cline{10-12} 
   &0 &10  & 14  &&0 &10  & 13&  &0 &10  & 24/28\\  \cline{2-4} \cline{6-8} \cline{10-12} 
  &10 &10 & 11 &&10 &10  &10 &  &10 &10  & 10/19\\  \cline{2-4} \cline{6-8} \cline{10-12} 
   &-10 &10  & 12   &&-10 &10  & 15 &  &-10 &10  & 15/19 \\ \cline{2-4} \cline{6-8} \cline{10-12} 
 \hline\hline
\end{tabular}

}
\label{tabthreshold}
\end{center}
\end{table}

\subsection{Evaluation of Point Cloud Registration Accuracy} \label{test1}
\begin{figure*}
\centering
\includegraphics[width=17.5cm]{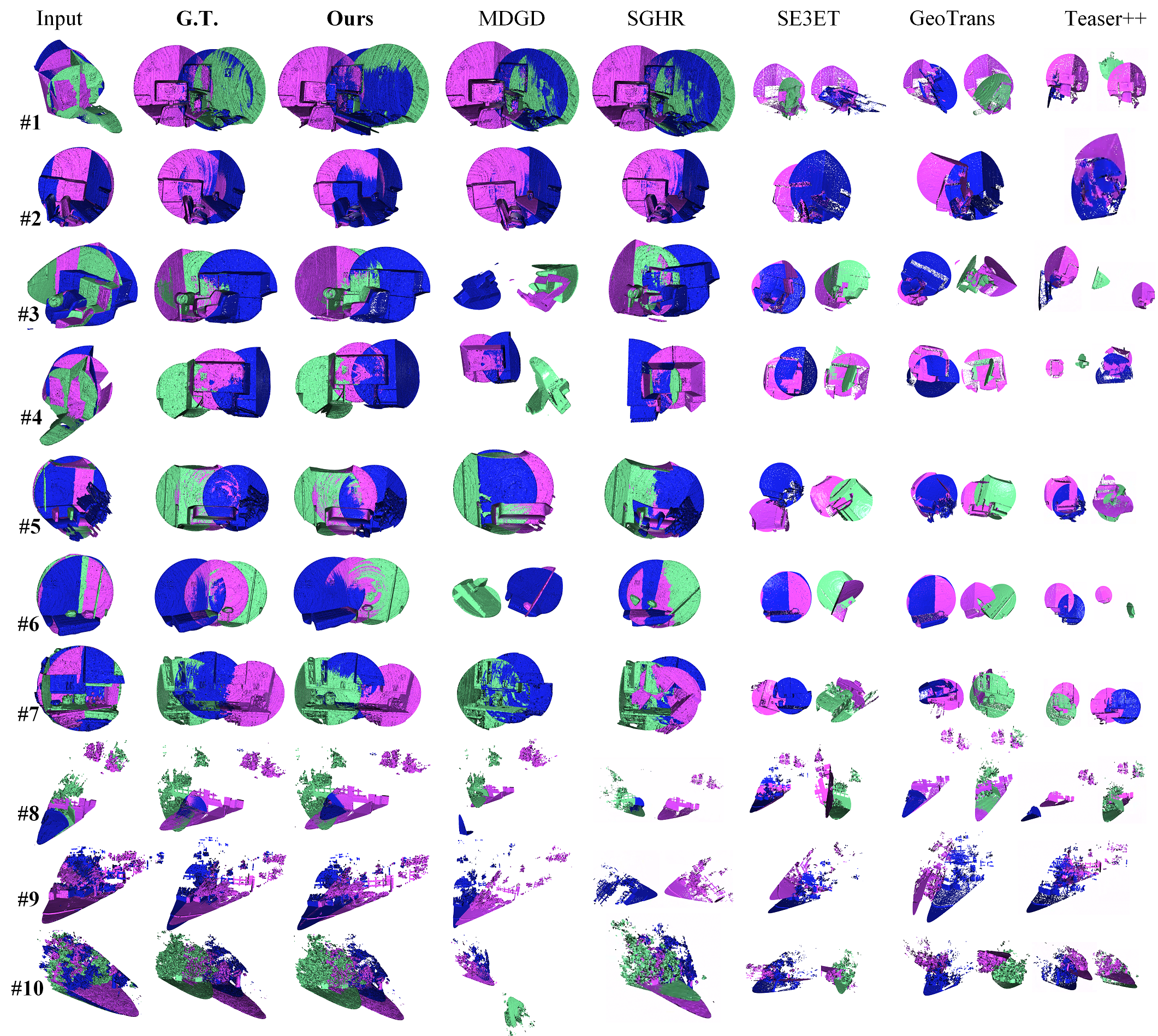}
{
\caption{A comparison with SOTA methods, including MDGD \cite{mdgd}, SGHR \cite{sghr}, SE3ET \cite{se3et}, GeoTrans \cite{geotransformer}, and Teaser++ \cite{teaser}, on ten scenes. The scenes include the office (1-3), the meeting room (4), the lounge (5,6), the kitchen (7), the office building (8,9), and the thicket (10). Each scene consists of three scans, except scenes 2 and 9, which are composed of two scans.} \label{fig8}
}
\end{figure*}

In this test, we evaluate the proposed method against competitors in terms of point cloud registration accuracy.
\\
\noindent\textbf{Data.}
Given that the existing point cloud registration benchmarks \cite{3dmatch,eth,scan} lack fiducial markers in the scenes, we construct a new dataset, as shown in Fig. \ref{fig8}, with the Livox MID-40. As illustrated in the caption of Fig. \ref{fig8}, our dataset covers various indoor and outdoor scenes. Indoors, multiple 16.4 cm × 16.4 cm AprilTags \cite{ap3} are positioned in the environment. Outdoors, multiple 69.2 cm × 69.2 cm ArUcos \cite{aruco} are placed in the environment. Note that since the LiDAR fiducial markers in this work are thin sheets of objects attached to other surfaces, they are invisible in the 3D point clouds. This is infeasible if we adopt LiDARTags \cite{lt} or calibration boards \cite{cal,cal2,a4} to provide fiducials. These scenes are challenging due to low overlap, and the overlap rate \cite{pre} of each scene is presented in Table \ref{tabnew}. The ground truth poses are obtained by manually registering through \textit{CloudCompare} \cite{cloudcompare}, a popular 3D point cloud processing software.
\\

\noindent\textbf{Competitors and Metrics.} Competitors include the latest state-of-the-art (SOTA) multiview point cloud registration methods, MDGD \cite{mdgd} (RA-L$^{\prime}$24), SGHR \cite{sghr} (CVPR$^{\prime}$23), and the SOTA pairwise methods, SE3ET \cite{se3et} (RA-L$^{\prime}$24), GeoTrans \cite{geotransformer} (TPAMI$^{\prime}$23), and Teaser++ \cite{teaser} (T-RO$^{\prime}$20). All the competitors are learning-based methods, except for Teaser++, which is a geometry-based method. For all pairwise methods, we manually select pairs of point clouds that have overlap as the inputs. We employ the root-mean-square errors (RMSEs) \cite{shuo} as the metric. We also compare the inference time of the methods with an AMD Ryzen 7 5800X CPU.
 \\
\begin{table*}[htb]
{
\caption{quantitative comparison with MDGD \cite{mdgd}, SGHR \cite{sghr}, SE3ET \cite{se3et}, GeoTrans \cite{geotransformer}, and Teaser++ \cite{teaser} on our dataset.}
\centering
	\resizebox{2.0\columnwidth}{!}{
\begin{tabular}{c|c|c|c|c|c|c|c|c|c|c|c}
\hline\hline

\multicolumn{2}{c|}{Scene $\#$ } & 1 (Office)  &2 (Office) & 3 (Office)  & 4 (Meeting Room) &5 (Lounge)  & 6 (Lounge)&  7 (Kitchen)& 8 (Building)&  9 (Building) & 10 (Thicket) \\ 

\multicolumn{2}{c|}{Avg/Min Overlap Rate (\%)} & 27.17/14.29  &46.88/46.88 &44.66/22.26 & 25.80/1.40 &43.82/15.48 & 50.66/27.24&  31.31/4.19 & 19.06/12.20 &  44.65/44.65 & 12.16/5.42 \\ 
\hline
\multirow{3}{*}{Teaser++ \cite{teaser}} &$\mathrm{RMSE}_{R}$ (rad) & 1.504  & 1.795 & 2.394 & 2.277 & 2.493&  1.946& 1.632&  1.818 & 1.779 & 1.876 \\
&$\mathrm{RMSE}_{T}$ (m) & 1.724 & 1.890 & 2.075 & 2.252& 1.911&  1.733& 2.089&  1.861 & 1.848 & 1.891 \\
(T-RO$^{\prime}$20)& Time (s) & 313.1 & 298.3 & 336.4 & 322.3& 388.9 &  419.8& 441.5&  397.1 & 327.2 & 401.2 \\
\hline
\multirow{3}{*}{GeoTrans \cite{geotransformer}} &$\mathrm{RMSE}_{R}$ (rad) & 1.331  & 1.451 & 1.935 & 1.632 & 1.532&  1.471& 1.198&  1.355 & 1.377 & 1.526 \\
&$\mathrm{RMSE}_{T}$ (m) & 1.502 & 1.671 & 1.912 & 1.552& 1.325&  1.235& 1.132&  1.730 & 1.554 & 1.531 \\
(TPAMI$^{\prime}$23)& Time (s) & 91.8 & 87.3 & 95.1 & 93.9& 105.3 &  122.3& 127.3&  110.5 & 91.1 & 121.4 \\
\hline
\multirow{3}{*}{SE3ET \cite{se3et}} &$\mathrm{RMSE}_{R}$ (rad) & 0.925  & 0.834 & 1.776 & 1.358 & 1.381&  1.352& 1.077&  1.156 & 1.232 & 1.376 \\
&$\mathrm{RMSE}_{T}$ (m) & 0.996 & 1.113 & 1.736 & 1.273& 1.207&  1.130& 1.095&  1.469 & 1.413 & 1.392 \\
(RA-L$^{\prime}$24)& Time (s) & 116.8 & 94.6 & 99.5 & 95.7& 115.7 &  136.0& 139.3&  123.5 & 97.3 & 132.0 \\
\hline
\multirow{3}{*}{SGHR \cite{sghr}} & $\mathrm{RMSE}_{R}$ (rad)  & 0.152  & \textbf{0.060}  & 1.198 & 1.825 & 1.311  &  1.696& 3.42 &  2.792 & 3.034 &2.810 \\
& $\mathrm{RMSE}_{T}$ (m) & 0.020 & 0.010 & 0.094 & 0.156& 0.231 &  0.274&  0.949&  2.171 & 0.686 & 1.787 \\
(CVPR$^{\prime}$23)& Time (s) & 846.1 & 785.0 & 886.5 & 825.3& 851.7&  979.3& 983.6&  909.9 & 794.6 &950.4 \\
\hline
\multirow{3}{*}{MDGD \cite{mdgd}} & $\mathrm{RMSE}_{R}$ (rad)  & \textbf{0.032}  & 0.062  & 1.457 & 0.933 & 0.679  &  1.715& 0.856 &  0.722 & 0.651 &0.779 \\
& $\mathrm{RMSE}_{T}$ (m) & \textbf{0.015} & \textbf{0.009} & 0.352 & 1.035& 0.113 &  0.422&  0.337&  1.553 & 0.686 & 0.939 \\
(RA-L$^{\prime}$24)& Time (s) & 626.1 & 573.3 & 645.5 & 619.2& 612.0&  726.4 & 730.5&  689.6 & 590.6 &717.4 \\
\hline
\multirow{3}{*}{\textbf{Ours}} & $\mathrm{RMSE}_{R}$ (rad) & 0.036 & 0.068 & \textbf{0.089} & \textbf{0.065}& \textbf{0.088}&  \textbf{0.078}& \textbf{0.067}&  \textbf{0.069} & \textbf{0.087} & \textbf{0.101} \\
& $\mathrm{RMSE}_{T}$ (m) &   0.017 &  0.011 &  \textbf{0.028} &  \textbf{0.031}& \textbf{ 0.032}&   \textbf{0.043}&  \textbf{0.019}&  \textbf{0.082} &  \textbf{0.069} & \textbf{0.077} \\
        
& Time (s) & \textbf{31.1}   & \textbf{21.2}  & \textbf{35.7}   & \textbf{32.6 }& \textbf{36.2} &  \textbf{43.5}& \textbf{53.8} &  \textbf{39.9} & \textbf{24.9}  &\textbf{42.2} \\

\hline\hline
\end{tabular}
}
\label{tabnew}
}
\end{table*}
\noindent\textbf{Results and Analysis.} The qualitative and quantitative results are presented in Fig. \ref{fig8} and Table \ref{tabnew}, respectively. 
The pairwise methods \cite{se3et,geotransformer,teaser}, although provided manually selected pairs of point clouds with overlap, struggle in all scenes.
This is because they are tailored for extracting geometric features either in a learning-based manner \cite{se3et,geotransformer} or through conventional geometry \cite{teaser}. Thus, when the overlap ratio between point clouds is low and the overlapped regions lack sufficient geometric features, these methods struggle to find features and utilize them to align point clouds. 
In contrast, the multiview point cloud registration methods \cite{mdgd,sghr} show some successful cases. In particular, SGHR \cite{sghr} successfully registers the point clouds in scenes 1 to 3 and one point cloud pair in scene 4. MDGD \cite{mdgd}, built upon the framework of SGHR \cite{sghr} but developing a new matching distance-based overlap estimation module, aligns scenes 1 and 2 with decent accuracy and also successfully registers one pair of point clouds in scenes 4, 5, 7, and 10. The reason is that, in addition to learning pairwise registration, the multiview point cloud registration methods \cite{mdgd,sghr} also learn to further optimize or refine the pose graph constructed using pairwise registration results.
Despite this, they fail in other scenes. Reviewing their success cases, it is found that these indoor scenes are similar to those in their training dataset \cite{3dmatch}. While failure cases, such as scenes 4 to 10, are rare or lacking in the training dataset. This indicates that their generalization to unseen scenarios is limited due to the out-of-distribution problem. The proposed method achieves the best performance. Specifically, our method does not require the overlapping regions to have explicit geometric features, thanks to the thin-sheet format of our LiDAR fiducial markers. Moreover, by virtue of the proposed adaptive marker detection algorithm, our method can robustly align any novel indoor or outdoor scenes with thin-sheet markers.
The proposed method also yields the best efficiency as it focuses on registering point clouds through LiDAR fiducial markers rather than analyzing the entire point clouds. The process of our method takes several dozen seconds, while other methods need several minutes on the AMD Ryzen 7 5800X CPU.
\begin{figure*}
	\centering
	\includegraphics[width=17.5cm]{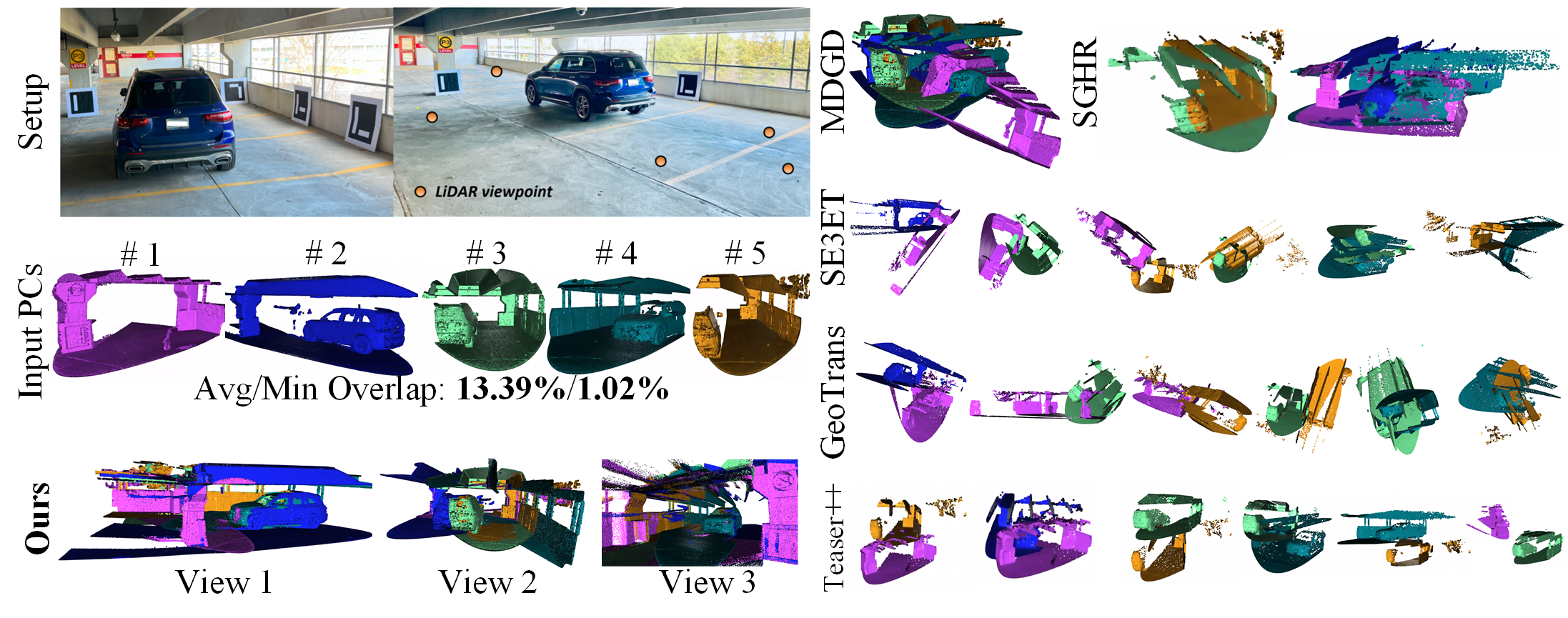}
	\caption{An illustration of the experimental setup and a visual comparison of the proposed method against the SOTA methods (MDGD \cite{mdgd}, SGHR \cite{sghr}, SE3ET \cite{se3et}, GeoTrans \cite{geotransformer}, and Teaser++ \cite{teaser}) regarding instance reconstruction.  }
	\label{glbtraj}
\end{figure*} 

\begin{figure}[thpb]
	\centering
	\includegraphics[width=3.3in]{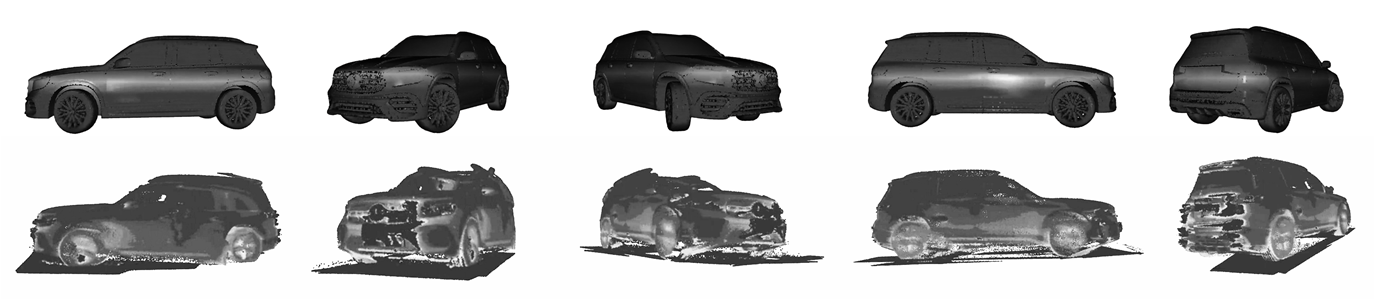}
	\caption{A visual comparison of the ground truth shape (top) and the reconstructed shape using the proposed method (bottom).}
	\label{glb}
\end{figure} 

\subsection{Application 1: 3D Asset Collection from Sparse Scans} \label{test2}
Collecting the complete shape of a novel object \cite{cd} from sparse observations is advantageous, as sparse scans require less labor and offer better efficiency. However, it is also a challenging task due to the low overlap between scans. In this test, we evaluate the instance reconstruction quality of the proposed method.
\\
\noindent\textbf{Data.} The experimental setup is depicted in Fig. \ref{glbtraj}. Four 69.2 cm $\times$ 69.2 cm ArUco \cite{aruco} markers are placed in the environment. We employ the Livox MID-40 LiDAR to scan the vehicle from five significantly different viewpoints. As shown in Fig. \ref{glbtraj}, this is also a challenging low overlap case, with the average and minimum overlap rates being 13.39\% and 1.02\%, respectively.
To evaluate the instance reconstruction quality, we use \textit{Supervisely} \cite{super}, a popular 3D annotation tool, to extract the point cloud of the vehicle from the reconstruction result.
Since the vehicle's make and model (Mercedes-Benz GLB) are known, a high-fidelity 3D model from a 3D assets website serves as the ground truth shape. \\
\noindent\textbf{Competitors and Metrics.} The competitors are the SOTA point cloud registration methods, including MDGD \cite{mdgd}, SGHR \cite{sghr}, SE3ET \cite{se3et}, GeoTrans \cite{geotransformer}, and Teaser++ \cite{teaser}. Pairs of point clouds with overlapping regions are manually provided to the pairwise methods \cite{se3et,geotransformer,teaser}. MDGD \cite{mdgd}, SGHR \cite{sghr}, and our method directly process the set of point clouds. In terms of metrics, we follow \cite{cd} and employ Chamfer Distance and Recall between the ground truth shape and the reconstructed shape to evaluate reconstruction quality.\\
\begin{table*}[htbp]
{
\caption{quantitative evaluation of the enhancement of SGHR \cite{sghr} and MDGD \cite{mdgd} due to the addition of Livox-3DMatch to the training.}
\centering
	\resizebox{2.0\columnwidth}{!}{
\begin{tabular}{c|c|c|c|c|c|c|c}
\hline\hline
\multirow{2}{*}{Method} & \multirow{2}{*}{Dataset} & RR $\uparrow$  & \multirow{2}{*}{Dataset} &RR $\uparrow$ & \multirow{2}{*}{Dataset} &  $\mathrm{RMSE}_{T}$ (m) $\downarrow$ & $\mathrm{RMSE}_{R}$ (deg) $\downarrow$ \\ 
  & & (\%) &  &(\%) &  &  Mean/Med & Mean/Med\\ \hline
SGHR \cite{sghr} & \multirow{4}{*}{3DMatch \cite{3dmatch}}  & 92.68 & \multirow{4}{*}{ETH \cite{eth}} &93.74 & \multirow{4}{*}{ScanNet \cite{scan}} &  0.66/0.51 & 23.50/22.08 \\ 
SGHR \cite{sghr} + \textbf{Livox-3DMatch (our data)} &  & \textbf{95.58} &  &\textbf{98.03} &  &  \textbf{0.51/0.45} & \textbf{20.87/17.21} \\ 
MDGD \cite{mdgd}  &  & 94.26 &  & 96.06 &  &  0.49/0.44 & 20.51/19.82 \\ 
MDGD \cite{mdgd} + \textbf{Livox-3DMatch (our data)} &  & \textbf{95.97} &  &\textbf{98.95} &  &  \textbf{0.38/0.31} & \textbf{18.91/17.36} \\ \hline

\hline\hline
\end{tabular}
}
\label{tabsghr}
}
\end{table*}
\noindent\textbf{Results and Analysis.} The qualitative results are shown in Fig. \ref{glbtraj} and Fig. \ref{glb}. As shown in Fig. \ref{glbtraj}, MDGD \cite{mdgd} and SGHR \cite{sghr} successfully align the third and fifth scans. In particular, the third and fifth scans have the highest overlap ratio of 59.33\% among all pairs as they are captured from similar perspectives. However, when dealing with point clouds that have lower overlap ratios, the competitors \cite{mdgd,sghr,se3et,geotransformer,teaser} struggle to register them. This comparison illustrates that, although these existing methods can handle some unseen scenarios with high overlap, they are not generalizable to novel low overlap cases.
Moreover, the failure of these existing methods in the instance reconstruction task implies that they are not suitable for efficient 3D asset collection. By contrast, the proposed method successfully registers the unordered multiview point clouds. Especially, as presented in Fig. \ref{glb}, our reconstructed shape preserves intricate details well compared to the ground truth shape. From a quantitative perspective, the Chamfer Distance and Recall of our reconstructed result are \textbf{0.003} and \textbf{96.22\%}, respectively. The decent performance in terms of reconstruction quality metrics \cite{cd} indicates that L-PR can serve as a convenient, efficient, and low-cost tool for collecting high-fidelity 3D assets using the LiDAR sensor. 
In particular, we consider the proposed method efficient, as it only requires four or five scans from dramatically changed viewpoints to reconstruct the complete shape.

\subsection{Application 2: Training Data Collection and Enhancement of Existing Learning-Based Methods} \label{testadd}
Training data is crucial for learning-based methods. Most existing methods \cite{sghr,mdgd} are trained on the 3DMatch dataset \cite{3dmatch}. Augmenting the training data is beneficial for improving the generalization ability of learning-based methods. However, the existing methods cannot be utilized for collecting training data in unseen scenes, as unseen scenes imply out-of-distribution cases and the limited effectiveness of the existing methods.
The proposed method can serve as a valuable tool for collecting training data. In this test, we demonstrate that training data collected using L-PR can enhance the performance of the SOTA methods across various benchmarks.
\\
\noindent\textbf{Data.}  
Using the proposed method, all the point clouds shown in Figs. \ref{fig8} and \ref{glbtraj}, comprising 11 scenes with 33 scans, are aligned and processed into the format required for training by MDGD \cite{mdgd} and SGHR \cite{sghr}. The newly collected data is named Livox-3DMatch. We train the models \cite{mdgd,sghr} from scratch using only 3DMatch and a combination of 3DMatch and Livox-3DMatch.
\\
\noindent\textbf{Benchmarks and Metrics.}
We compare the performance of the models \cite{mdgd,sghr} trained with and without our data on three popular benchmarks: 3DMatch \cite{3dmatch}, ETH \cite{eth}, and ScanNet \cite{scan}. Following \cite{sghr}, we use Registration Recall (RR) to evaluate performance on 3DMatch \cite{3dmatch} and ETH \cite{eth}, and RMSEs to report performance on ScanNet \cite{scan}. 
\\
\noindent\textbf{Results and Analysis.} 
First, the number of pairs in the training data increases from 14,400 to \textbf{17,700}, a boost of \textbf{22.91\%}, thanks to the introduction of our data. 
The quantitative results are reported in Table \ref{tabsghr}. 
As seen, the RR of SGHR \cite{sghr} is increased by \textbf{2.90\%} on 3DMatch \cite{3dmatch} and \textbf{4.29\%} on ETH \cite{eth}. The translation error and rotation error on ScanNet \cite{scan} are decreased by \textbf{22.72\%} and \textbf{11.19\%}, respectively.
The RR of MDGD \cite{mdgd} is improved by \textbf{1.71\%} on 3DMatch \cite{3dmatch} and \textbf{2.89\%} on ETH \cite{eth}. The translation error and rotation error on ScanNet \cite{scan} are decreased by \textbf{22.45\%} and \textbf{7.80\%}, respectively.
The reason behind these improvements is mentioned in Section \ref{secfour}: the introduction of our data enriches the learnable features during training, which benefits the generalizability of the models.
\begin{figure*}
	\centering
	 \includegraphics[width=17.5cm]{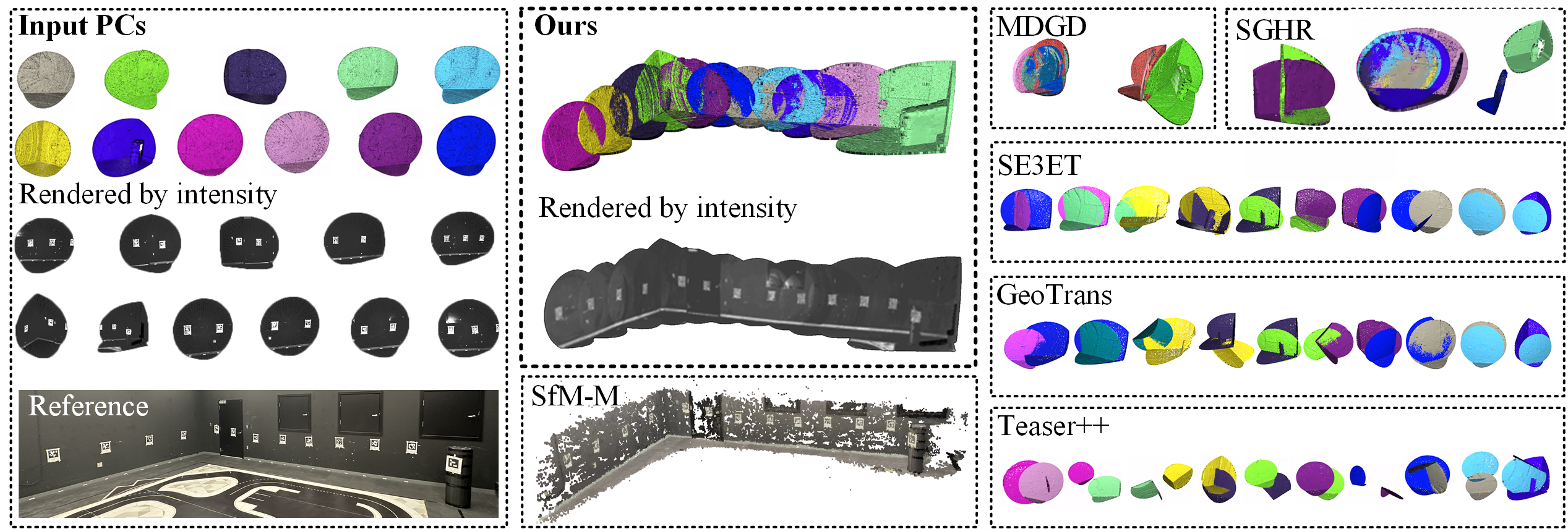}
	\caption{An illustration of the experimental setup and a visual comparison of the proposed method against the SOTA methods (MDGD \cite{mdgd}, SGHR \cite{sghr}, SE3ET \cite{se3et}, GeoTrans \cite{geotransformer}, Teaser++ \cite{teaser}, and SfM-M \cite{qingdao}) in a degraded scene.}
	\label{labpic}

\end{figure*}

\begin{figure}[ht]
	\centering
	\includegraphics[width=2.0in]{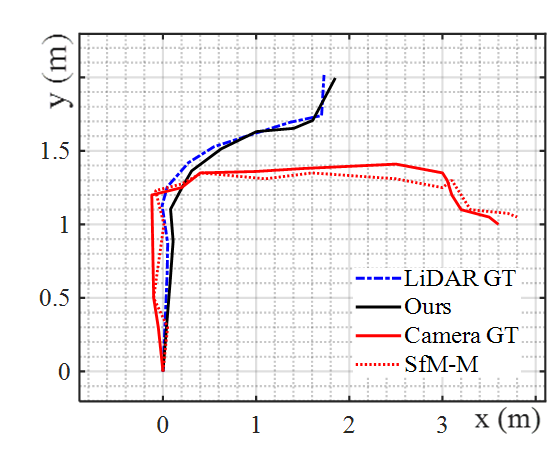}
	\caption{ A comparison of the sensor trajectories obtained from the proposed method and SfM-M \cite{qingdao}. G.T. refers to the ground truth. }
	\label{labtraj}
 
\end{figure}

\subsection{Application 3: Reconstructing a Degraded Scene} \label{test3}
An important application of fiducial markers in the real world is enhancing the robustness of reconstruction and localization in degraded scenes when additional sensors are unavailable. Therefore, in this test, we evaluate our method in a textureless, degraded scene. 
\\
\noindent\textbf{Data.} The setup is shown in the reference of Fig. \ref{labpic}. This scenario has repetitive structures and weak geometric features. We attach thirteen 16.4 cm $\times$ 16.4 cm AprilTags to the wall. The LiDAR scans the scene from 11 viewpoints. We also captured 72 images with an iPhone 13 to use as input for SfM-M \cite{qingdao}. The ground truth trajectories are given by an OptiTrack Motion Capture system. 
\\
\noindent\textbf{Competitors and Metrics.} 
The competitors are the SOTA point cloud registration methods \cite{mdgd,sghr,se3et,geotransformer,teaser}. However, considering these competitors struggle in the degraded scene, we add SfM-M \cite{qingdao}, the SOTA marker-based SfM method, as a competitor. This addition enables a comprehensive and meaningful comparison. We employ the RMSEs as the metric.
\\
\noindent\textbf{Results and Analysis.} The qualitative results are presented in Fig. \ref{labpic}. As seen, none of the point cloud registration methods \cite{mdgd,sghr,se3et,geotransformer,teaser} can align a single pair of point clouds. 
This is because these cases only have planar overlap regions and thus are even more challenging than those in Fig. \ref{fig8} and Fig. \ref{glbtraj}. 
In comparison, the proposed method and SfM-M \cite{qingdao} successfully reconstruct this degraded scene as they utilize the fiducial markers. The comparison of these two methods in terms of sensor trajectories is shown in Fig. \ref{labtraj}. The quantitative results shown in Table \ref{427} demonstrate that the proposed approach achieves better localization accuracy, which is expected given that LiDAR is a ranging sensor. 
\begin{table}[h]
\caption{comparison with SfM-M \cite{qingdao} regarding localization accuracy in a degraded scene. }
	\centering
	\resizebox{0.8\columnwidth}{!}{
		\begin{tabular}{c|c|c}
			\hline\hline
				Method $\backslash$ Metric & $\mathrm{RMSE}_{T}$ (m) $\downarrow$ & $\mathrm{RMSE}_{R}$ (rad) $\downarrow$\\ \hline
   SfM-M \cite{qingdao} & 0.055 & 0.049  \\ \hline
   \textbf{Ours} & 0.049 & 0.038
\\ \hline  \hline
			
		\end{tabular}
	}
		\label{427}
\end{table}
\subsection{Application 4: Localization in a GPS-denied Environment} \label{test4}
\begin{figure}[ht]
	\centering
	\includegraphics[width=3.3in]{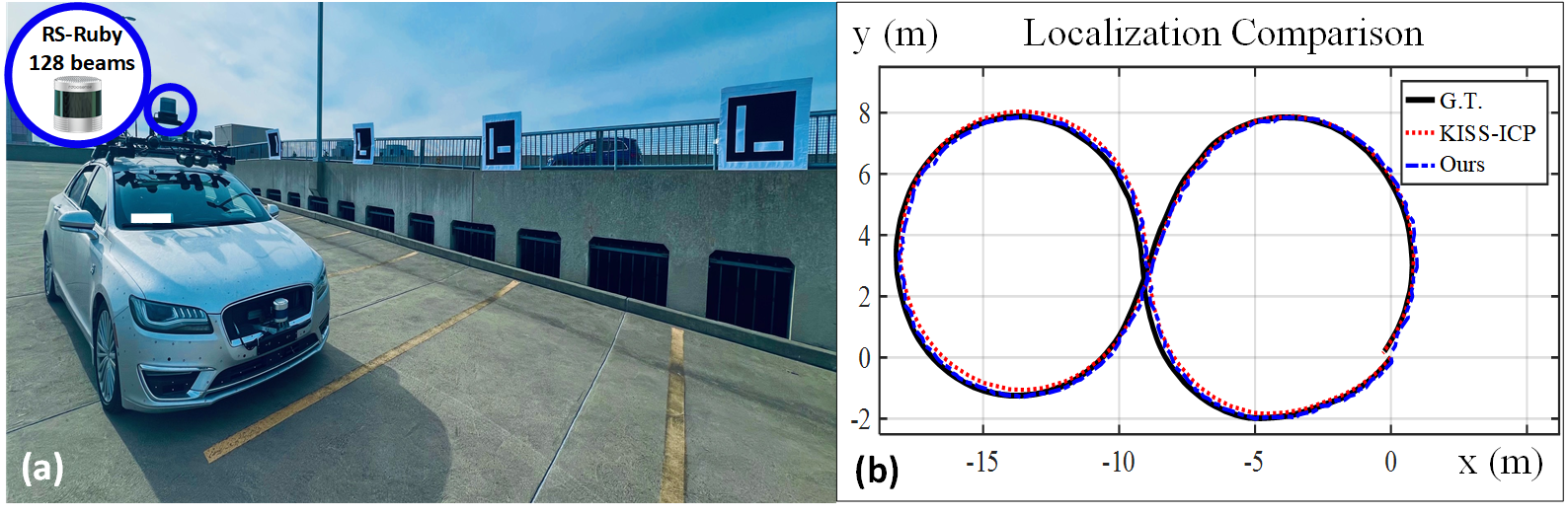}
 
	\caption{ (a): An illustration of the experimental setup. (b): A comparison of the trajectories given by the proposed method and KISS-ICP \cite{kiss}. G.T. refers to the ground truth. }
	\label{rob}
 
\end{figure}

Localization is also a crucial application of point cloud registration methods. Fiducial markers are a popular tool for providing localization information in GPS-denied environments, such as indoor parking lots. In this test, we evaluate the proposed method in this context.
\\
\noindent\textbf{Data.} The experimental setup is shown in Fig. \ref{rob}(a): four 69.2 cm $\times$ 69.2 cm ArUco \cite{aruco} markers are deployed in the environment. The vehicle, equipped with an RS-Ruby 128-beam mechanical LiDAR, follows an 8-shaped trajectory without pausing and samples 364 LiDAR scans. We conduct the experiment on the roof of a large parking lot to acquire the ground truth trajectory from the Real-Time Kinematic. 
\\
\noindent\textbf{Competitors and Metrics.}
We compare the proposed method with KISS-ICP \cite{kiss} in terms of localization accuracy. We employ the RMSEs
as the metric. 
\\
\noindent\textbf{Results and Analysis.} 
The visual comparison and quantitative comparison of localization results are presented in Fig. \ref{rob}(b) and Table \ref{outtab}, respectively. Our method exhibits less drift in the middle of the trajectory and demonstrates better overall localization accuracy.
\begin{table}[th]
\caption{
Comparison of localization results between KISS-ICP \cite{kiss} and our method.
}
	\centering
	\resizebox{0.8\columnwidth}{!}{
		\begin{tabular}{c|c|c}
			\hline\hline
				Method $\backslash$ Metric & $\mathrm{RMSE}_{T}$ (m) $\downarrow$ & $\mathrm{RMSE}_{R}$ (rad) $\downarrow$\\ \hline
   KISS-ICP \cite{kiss} & 0.198 & 0.162  \\ \hline
   \textbf{Ours} & 0.172 & 0.139
\\ \hline  \hline
			
		\end{tabular}
	}
		\label{outtab}
\end{table}
\subsection{Application 5: 3D Map Merging} \label{test5}
\begin{figure}[h]
	\centering
	\includegraphics[width=3.3in]{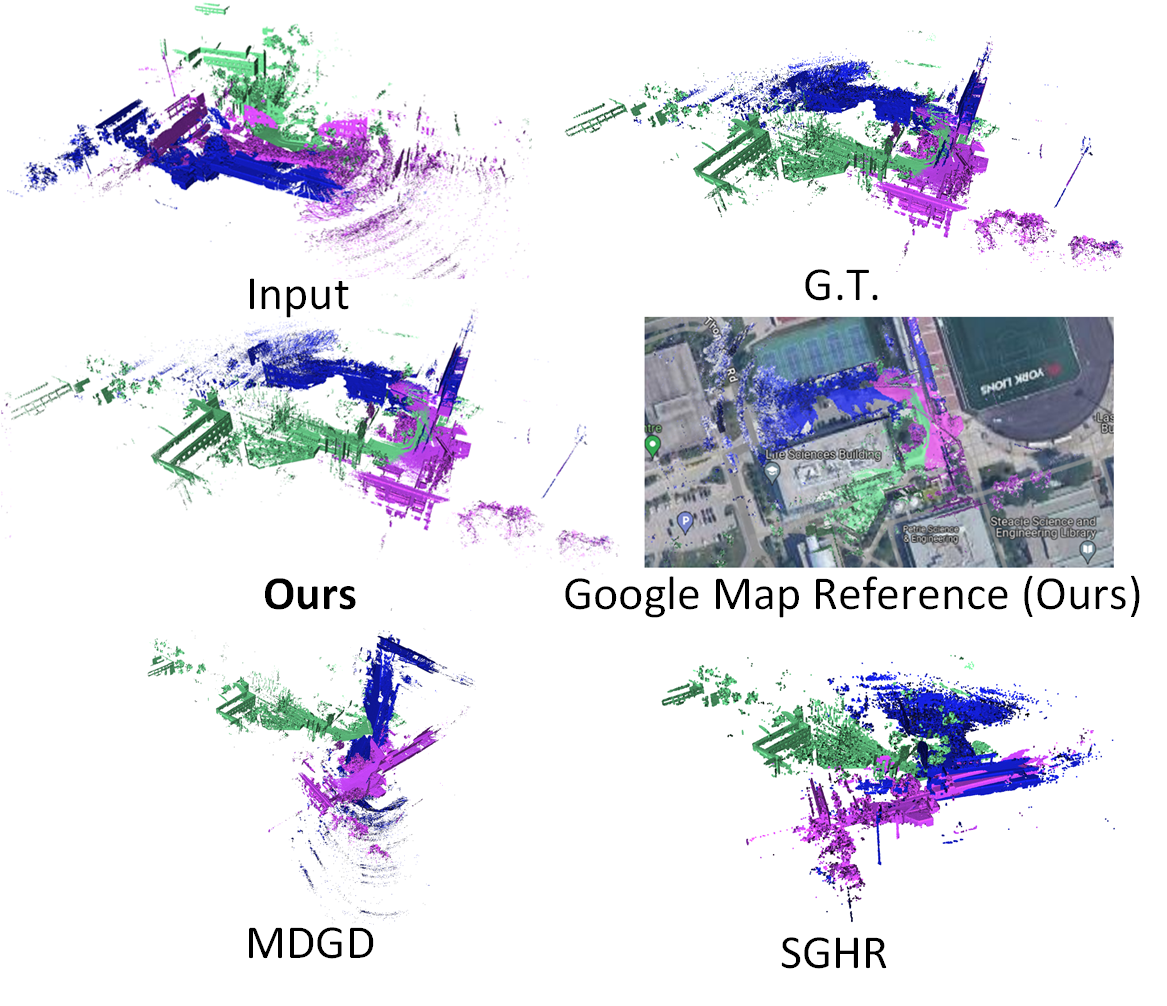}
	\caption{A comparison of the 3D map merging results of MDGD \cite{mdgd}, SGHR \cite{sghr}, and our method. }
	\label{merge}

\end{figure}

To validate the performance of the proposed method in large-scale scenarios, we apply L-PR to the 3D map merging task in this test, which involves merging multiple large-scale, low overlap 3D maps into a single frame.
\\
\noindent\textbf{Data.} We collected three large-scale LiDAR maps. They are constructed using the SOTA LiDAR-based SLAM method, Traj-LO \cite{traj}, by scanning the York University campus with a Livox MID-40 LiDAR. The ground truth poses between the maps are manually obtained using \textit{CloudCompare} \cite{cloudcompare}. 
\\
\noindent\textbf{Competitors and Metrics.}
The competitors are SOTA multiview point cloud registration methods, including MDGD \cite{mdgd} and SGHR \cite{sghr}. Moreover, unlike previous tests, the LiDAR fiducial markers on the 3D maps are localized using the algorithm from our previous work \cite{mapmerge}, and the marker detection results serve as input for the L-PR pipeline.
We employ the RMSEs as the metric.
\\
\noindent\textbf{Results and Analysis.}
The visual and quantitative comparisons are shown in Fig. \ref{merge} and Table \ref{tabmap}, respectively. As shown in Fig. \ref{merge}, neither MDGD nor SGHR can address this challenging task. This is due to the fact that the overlap in large-scale maps is too scarce. Specifically, the overlap rates \cite{pre} are 4.87\% between map 1 and map 2, 3.96\% between map 1 and map 3, and 2.59\% between map 2 and map 3. On the other hand, both MDGD and SGHR start by analyzing the features of each individual point cloud. As the scale of the point cloud becomes larger, the portion of the features belonging to the overlap regions decreases, making them unsuitable for large-scale, low overlap map merging. In addition, as the scale of point clouds becomes larger, the absolute error values of MDGD and SGHR also increase. In comparison, even though the scales of the point clouds in this test are much larger than those in the previous test, the proposed method successfully merges these large-scale, low overlap 3D maps. This stems from the observation that the proposed method focuses on utilizing the thin-sheet LiDAR fiducial markers and is not sensitive to scale changes. As introduced in \cite{mapmerge}, the accuracy of marker localization degrades as the scale increases. Thus, the absolute error values of our method also increase compared to previous tests.

\begin{table}[h]
\caption{Comparison of 3D map merging results between MDGD \cite{mdgd}, SGHR \cite{sghr}, and our method.}
	\centering
	\resizebox{0.8\columnwidth}{!}{
		\begin{tabular}{c|c|c}
			\hline\hline
				Method $\backslash$ Metric & $\mathrm{RMSE}_{T}$ (m) $\downarrow$ & $\mathrm{RMSE}_{R}$ (rad) $\downarrow$\\ \hline
   MDGD \cite{mdgd} & 2.7883 & 1.9210  \\ \hline
   SGHR \cite{sghr} & 4.0933 & 2.3925  \\ \hline
   \textbf{Ours} & 0.2305 & 0.1771
\\ \hline  \hline
			
		\end{tabular}
	}
		\label{tabmap}
\end{table}

\subsection{Ablation Studies} \label{ab}
\begin{figure}[htbp]
	\centering
	\includegraphics[width=3.3in]{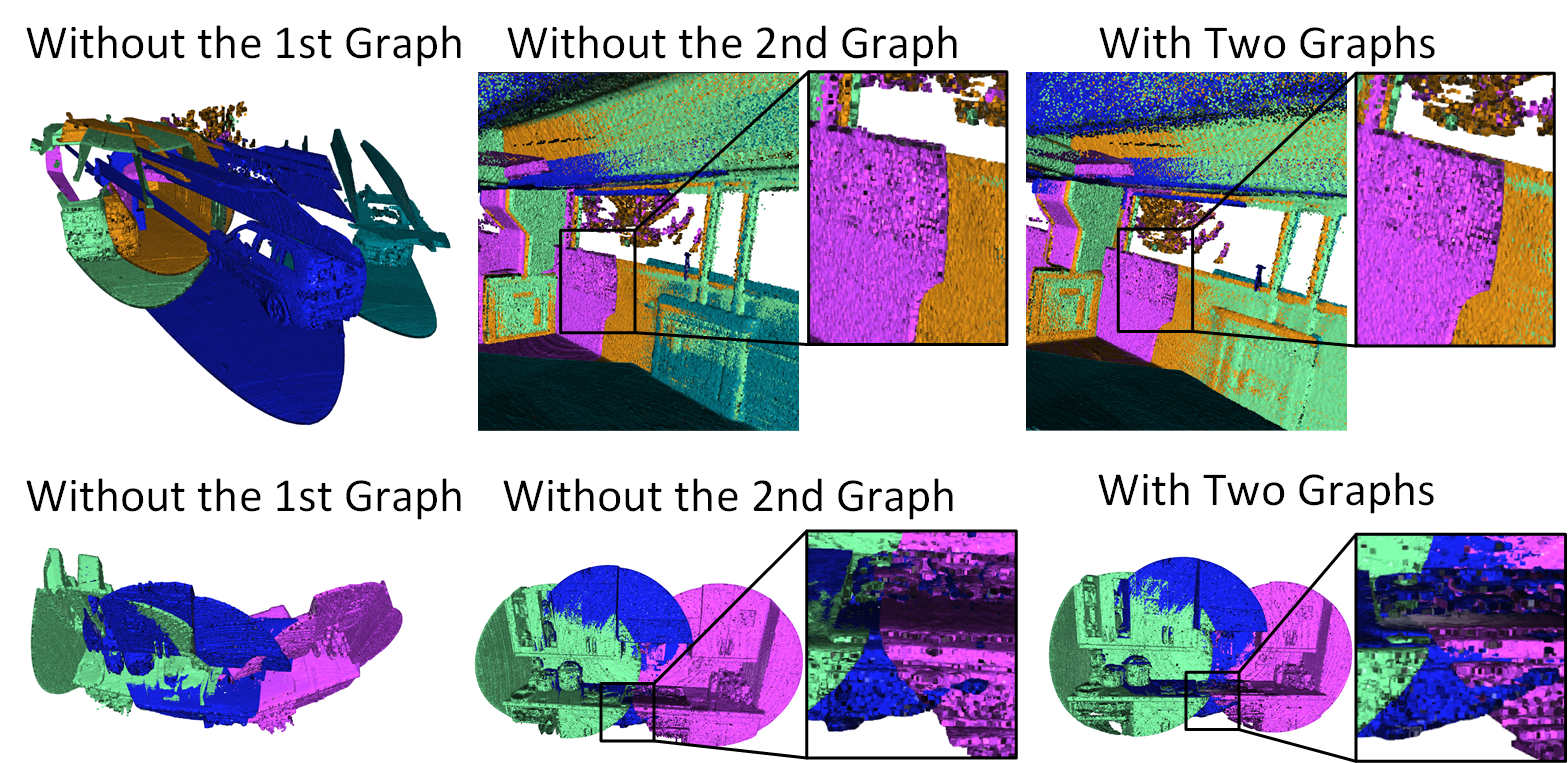}
	\caption{A comparison of the point cloud registration results without the first graph, without the second graph, and with both graphs for two cases.}\label{abstudy}
\end{figure}

The proposed L-PR framework is composed of two levels of graphs. To demonstrate the necessity of this overall architecture, we conduct ablation studies in this section. In particular, we study the effects of removing the first and second graphs in two cases: Fig. \ref{glbtraj} and the kitchen scene in Fig. \ref{fig8}. The visual comparison and quantitative results are shown in Fig. \ref{abstudy} and Table \ref{abtab}, respectively. When the first graph is removed, the factors representing the relative poses between frames in Fig. \ref{second} have to be set to identity due to the lack of initial values provided by the first graph. 
However, the role of the second graph is to further optimize the variables based on their initial values rather than finding the optimal solution from scratch.
Consequently, as shown in Fig. \ref{abstudy} and Table \ref{abtab}, severe misalignment or degradation in registration accuracy occurs.
When the second graph is removed, the multiview point clouds are directly registered using the initial values obtained from the first graph, without any further refinement. 
As shown in Fig. \ref{abstudy} and Table \ref{abtab}, removing the second graph causes a slight degradation in registration accuracy.
Moreover, the degradation in the kitchen case of Fig. \ref{fig8} is slighter than that in Fig. \ref{glbtraj}.
This is because the effect of the second graph is case-by-case, determined by the quality of the initial values provided by the first graph. 
Namely, the better the initial values are, the less important the second graph becomes. However, in the real world, pose estimation of markers cannot be perfect. In addition, as seen in Fig. \ref{second}, we also add the marker corners to the second graph so that the pose estimation of each individual marker can be further optimized along with other variables in the graph.
Therefore, it is beneficial to apply the second graph in practice. In summary, L-PR adopts a coarse-to-fine pipeline, where the first graph corresponds to the coarse stage, while the second graph is the fine stage. Removing any of them will cause performance degradation.

\begin{table}[h]
\caption{Ablation studies on the first and second graphs.}
	\centering
	\resizebox{1.0\columnwidth}{!}{
		\begin{tabular}{c|c|c|c}
			\hline\hline
				 Scene & Framework & $\mathrm{RMSE}_{T}$ (m) $\downarrow$ & $\mathrm{RMSE}_{R}$ (rad) $\downarrow$\\ \hline
\multirow{3}{*}{Fig. \ref{glbtraj}} & without first graph & 0.431 & 0.459 \\
& without second graph & 0.078 & 0.084 \\
& with both graphs & 0.066 & 0.072 \\
\hline
\multirow{3}{*}{Fig. \ref{fig8}} & without first graph & 0.088 & 0.112 \\
& without second graph & 0.021 & 0.073 \\
(Kitchen) & with both graphs & 0.019 & 0.067 \\
\hline  \hline
			
		\end{tabular}
	}
		\label{abtab}
\end{table}

\subsection{Limitations}
Despite the promising results of L-PR, there are potential limitations. Firstly, although LiDARs are robust to unideal illumination conditions, adverse weather phenomena such as rain, snow, and fog can affect LiDAR measurements, thereby reducing the effectiveness of L-PR. Secondly, the adopted low-cost LFMs, made of thin-sheet paper or boards, may deform after long-term use in the wild. However, this is not a concern for one-time applications such as data collection. Finally, deploying the LFMs requires some labor, but their value is demonstrated in this paper.

\section{Conclusion}
In this work, we develop L-PR, a novel framework that exploits LiDAR fiducial markers for unordered low overlap multiview point cloud registration. The markers in this work are thin sheets of paper attached to other surfaces, without impacting the 3D environment. L-PR consists of an adaptive threshold marker detection method and two levels of graphs.
The adaptive threshold method robustly detects markers even when there are dramatic changes in viewpoints among scans, thus overcoming the limitation of \cite{iilfm}.
The proposed two-level graph efficiently aligns the low overlap multiview point clouds directly from an unordered set by optimizing a MAP problem. 
We conduct qualitative and quantitative experiments to demonstrate the superiority of L-PR over the state-of-the-art competitors and the diverse applications of L-PR, including 3D asset collection, training data collection, reconstruction of a degraded scene, localization in a GPS-denied environment, and 3D map merging.
In particular, we collect a new training dataset named Livox-3DMatch using L-PR.
The Livox-3DMatch dataset expands the original 3DMatch training data from 14,400 pairs to 17,700 pairs, representing a 22.91\% increase. 
Training on this augmented dataset improves the performance of the SOTA learning-based methods on various benchmarks. In particular, the RR of SGHR \cite{sghr} is increased by 2.90\% on 3DMatch \cite{3dmatch} and 4.29\% on ETH \cite{eth}. The translation error and rotation error on ScanNet \cite{scan} are decreased by 22.72\% and 11.19\%, respectively. The RR of MDGD \cite{mdgd} increases by 1.71\% on 3DMatch \cite{3dmatch} and 2.89\% on ETH \cite{eth}. The translation error and rotation error on ScanNet \cite{scan} decrease by 22.45\% and 7.80\%, respectively. Future work includes utilizing L-PR to collect more 3D assets and point cloud registration training data to benefit the development of learning-based methods \cite{mdgd,sghr,guile}.
\bibliographystyle{IEEEtran} 
\bibliography{reference} 

\begin{IEEEbiography}[{\includegraphics[width=1in,height=1.25in,clip,keepaspectratio]{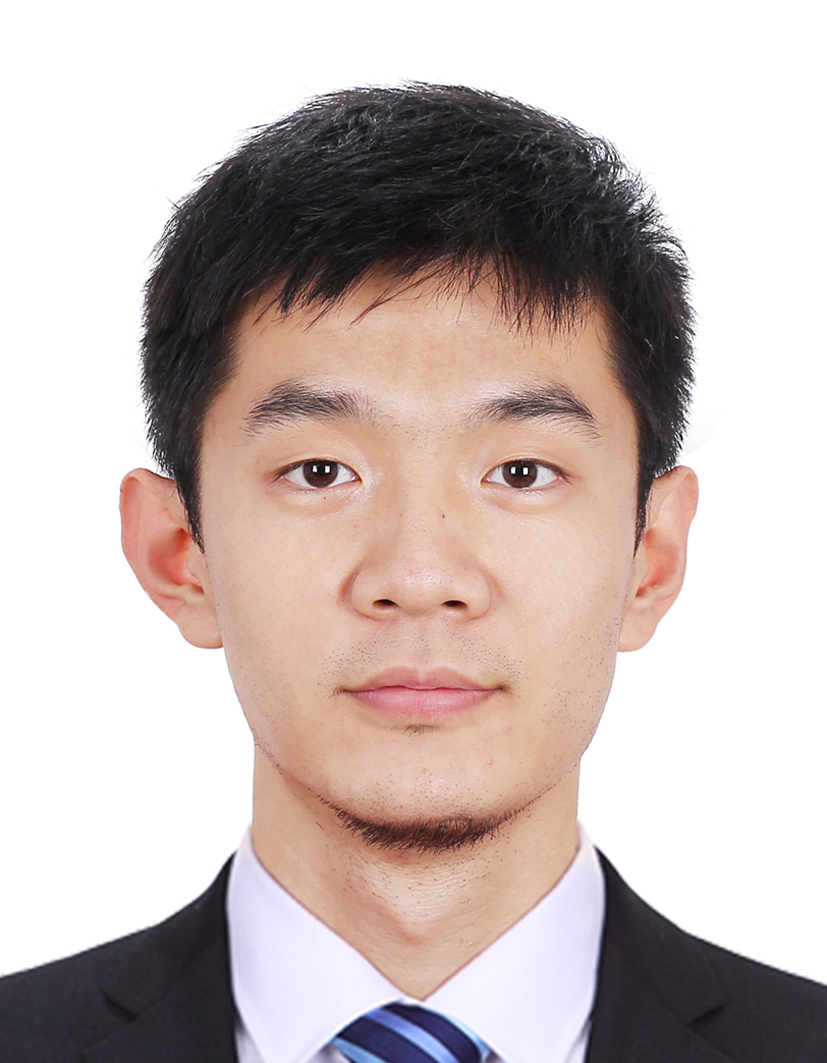}}] {Yibo Liu} (Member, IEEE) received the B.Sc. and M.Sc. degrees from Beihang University, Beijing, China, in 2017 and 2020, respectively. Since January 2020, he has been pursuing his Ph.D. degree in Earth and Space Science at York University, Toronto, Ontario, Canada. His research interests include Robot Vision and 3D Computer Vision.
\end{IEEEbiography}

\begin{IEEEbiography}[{\includegraphics[width=1in,height=1.25in,clip,keepaspectratio]{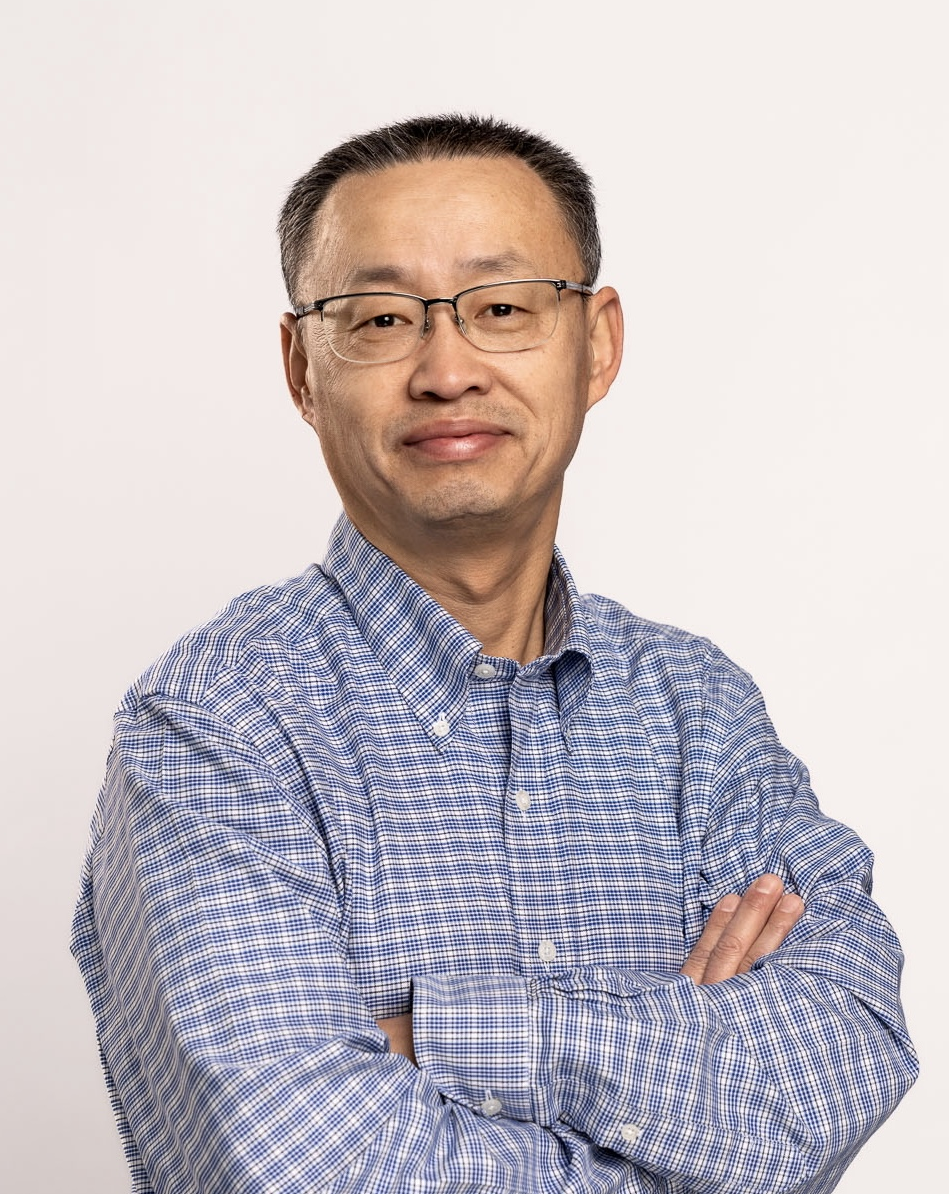}}]{Jinjun Shan} (Senior Member, IEEE) received the Ph.D. degree in spacecraft design from the Harbin Institute of Technology, Harbin, China, in 2002. He is currently a Full Professor of Space Engineering at the Department of Earth and Space Science and Engineering, York University, Toronto, ON, Canada. His research interests include dynamics, control, and navigation. Dr. Shan is a Fellow of Canadian Academy of Engineering (CAE), a Fellow of Engineering Institute of Canada (EIC) and a Fellow of American Astronautical Society (AAS). 
\end{IEEEbiography}

\begin{IEEEbiography}[{\includegraphics[width=1in,height=1.25in,clip,keepaspectratio]{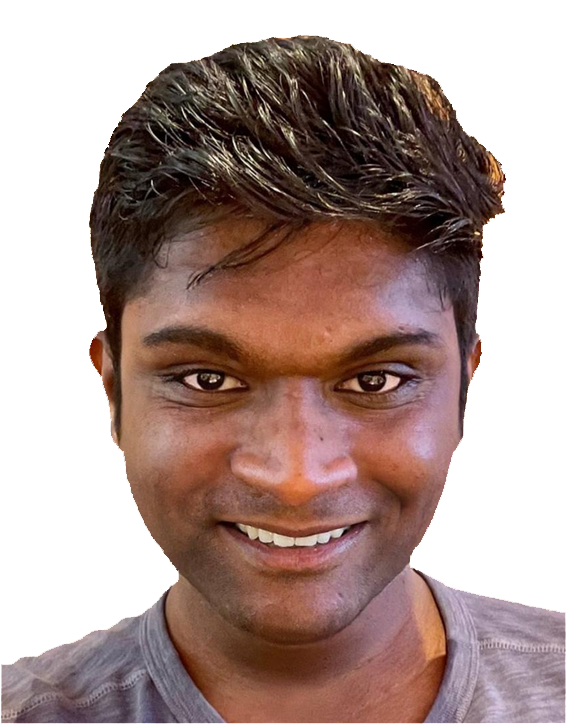}}]{Amaldev Haridevan}  received his B.Eng degree from York University, Toronto, Ontario, Canada in 2021. He is now working towards a M.Sc. degree with the Department of Earth and Space Science and Engineering, York University, Toronto, Ontario, Canada. His research interests are in Robot Vision and Reinforcement Learning.
\end{IEEEbiography}

\begin{IEEEbiography}[{\includegraphics[width=1in,height=1.25in,clip,keepaspectratio]{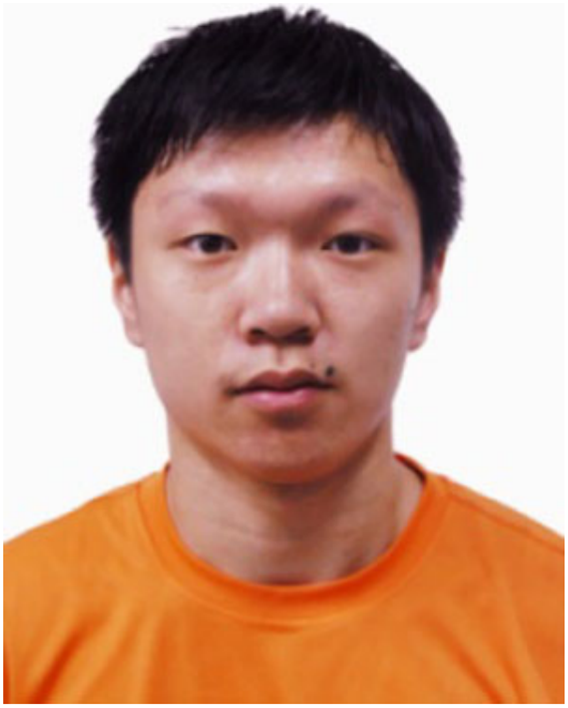}}]{Shuo Zhang} received the Ph.D. degree in precision instrument from Beihang University, Beijing, China, in 2016. He worked as a Postdoctoral Fellow with Tsinghua University, Beijing, China, from 2016 to 2019. He was a Postdoctoral Fellow with University of Alberta, Edmonton, AB, Canada, from 2019 to 2020. He is currently a Postdoctoral Fellow with the Department of Earth and Space Science and Engineering, York University, Toronto, ON, Canada. His research interests include optimal estimation, multi-sensor fusion, and navigation. 
\end{IEEEbiography}

\end{document}